\documentclass{article}

\usepackage{amsthm,amssymb}

\usepackage{algorithm,algcompatible,amsmath}
\DeclareMathOperator*{\argmax}{\arg\!\max}
\algnewcommand\INPUT{\item[\textbf{Input:}]}%
\algnewcommand\OUTPUT{\item[\textbf{Output:}]}%
\usepackage[utf8]{inputenc}
\usepackage{amsmath}
\usepackage{amssymb}
\usepackage{natbib}
\usepackage{enumitem}
\usepackage{pdfpages}
\usepackage{blkarray}
\usepackage{url}
\usepackage{mathtools}
\usepackage{geometry}
\usepackage{float}
\def\bmu{\mbox{\boldmath{$\mu$}}}

\newcommand{\bGamma}{\mbox{\boldmath{${\Gamma}$}}}

\newcommand{\balpha}{\mbox{\boldmath{${\alpha}$}}}
\newcommand{\btheta}{\mbox{\boldmath{${\theta}$}}}

\newcommand{\bLambda}{\mbox{\boldmath{${\Lambda}$}}}

\def\bSigma{\mbox{\boldmath{$\Sigma$}}}

\def\smt{{\mbox{\tiny T}}}

\def\bX{\mathbf X}
\def\bx{{\bf x}}

\def\bH{\mathbf H}
\def\bh{{\bf h}}

\def\bSb{\mathbf{S}_{b}}
\def\bSw{\mathbf{S}_{w}}

\def\bN{\mathbf N}

\def\bA{\mathbf A}

\def\bW{\mathbf W}

\def\bX{\mathbf X}
\def\bx{\mathbf x}
\def\bI{\mathbf I}

\def\bR{\mathbf R}
\def\bS{\mathbf S}

\def\bV{{\bf V}}

\def\bD{\mathbf D}

\def\rb{\textcolor[rgb]{0,0,1}}
\usepackage{color}
\usepackage{comment}
\usepackage{ulem}
\usepackage{soul}

\newtheorem{thm}{Theorem}

\def\spacingset#1{\renewcommand{\baselinestretch}%
{#1}\small\normalsize} \spacingset{1}

\title{Deep IDA: A Deep Learning Method for Integrative Discriminant Analysis of Multi-View Data with Feature Ranking--An Application to COVID-19 severity
}
\author{Jiuzhou Wang, Sandra E. Safo* \\ Division of Biostatistics\\University of Minnesota, MN }
\date{}

\usepackage{graphicx}

\begin{document}

\maketitle

\noindent\textbf{Abstract}
\vspace{3mm}
\\
COVID-19 severity is due to complications from SARS-Cov-2 but the clinical course of the infection varies for individuals, emphasizing the need to better understand the disease at the molecular level.  We use clinical and multiple molecular data (or views) obtained from patients with and without COVID-19 who were (or not) admitted to the intensive care unit to shed light on COVID-19 severity. Methods for jointly associating the views and separating the COVID-19 groups (i.e., one-step methods) have focused on linear relationships. The relationships between the views and COVID-19 patient groups, however, are too complex to be understood solely by linear methods. Existing nonlinear one-step methods cannot be used to identify signatures to aid in our understanding of the complexity of the disease. We propose Deep IDA (Integrative Discriminant Analysis) to address analytical challenges in our problem of interest. Deep IDA learns nonlinear projections of two or more views that maximally associate the views and separate the classes in each view, and permits feature ranking for interpretable findings.  Our applications demonstrate that Deep IDA has competitive classification rates compared to other state-of-the-art methods and is able to identify molecular signatures that facilitate an understanding of COVID-19 severity.\\

\noindent \textbf{Contact}: ssafo@umn.edu \\
\textbf{Supplementary Material:} Supplementary data  are available  online.\\

\newpage
\spacingset{1} 


\clearpage
\section{Introduction}
COVID-19 severity is  due to complications from the severe acute respiratory syndrome coronavirus 2 (SARS-CoV-2) 
but the clinical course of the infection varies for individuals. Research suggests that patients with and without severe COVID-19 have different genetic, pathological, and clinical signatures \citep{geneticscovid,overmyer2021large}. Further, beyond viral factors, COVID-19 severity depends on host factors, emphasizing the need to use molecular data to better understand the individual response of the disease \citep{overmyer2021large}. In \cite{overmyer2021large}, blood samples from patients admitted to the Albany Medical Center, NY from 6 April 2020 to 1 May 2020 for moderate to severe respiratory issues who had COVID-19 or exhibited COVID-19-like symptoms were collected and quantified for transcripts, proteins, metabolomic features and lipids. In addition to the molecular (or omics) data, several clinical and demographic data were collected at the time of enrollment. 
The authors analyzed each omics data separately, correlated the biomolecules with several clinical outcomes including disease status and severity, and also considered pairwise associations of the omics data to better understand  COVID-19 mechanisms. Their findings suggested that COVID-19 severity is likely due to dysregulation in lipid transport system. In this paper, we take a  holistic approach to integrate the omics data and the outcome (i.e., COVID-19 patient groups). In particular, instead of assessing pairwise associations and using unsupervised statistical methods as was done in \citep{overmyer2021large} to correlate the omics data, we  model the overall dependency structure among the omics data while simultaneously modeling the separation of the COVID-19 patient groups. Ultimately, our goal is to elucidate the molecular architecture of COVID-19 by identifying molecular signatures with potential to  discriminate  patients with and without COVID  who were or were not admitted to the the intensive care unit (ICU). 

There exists many linear (e.g., canonical correlation analysis, CCA [\cite{Hotelling:1936,Carroll:1968,SafoBIOM2017,safo2018sparse}], co-inertia analysis \citep{min2019penalized}) and  nonlinear (e.g.,\cite{Akaho:2001,Andrew:2013,Kan:2016,Benton2:2019} ) methods that could be used to associate the multiple views. Canonical Correlation Analysis with deep neural network (Deep CCA) \citep{Andrew:2013}, and its variations (e.g.\cite{Wang:2015},\citep{Benton2:2019}), for instance,  have been proposed to learn nonlinear projections of two or more views that are maximally correlated. Refer to \cite{Guo:2021} for a review of some CCA methods. These association-based methods are all unsupervised and they do not use the outcome data (i.e., class labels) when learning the low-dimensional  representations. 
A naive way of using the class labels and all the views simultaneously is to stack the different views and then to perform classification on the stacked data, but this approach does not appropriately model the dependency structure among the views. 

To overcome the aforementioned limitations, one-step linear methods (e.g., \cite{safosida:2021,zhang2018joint,Luo2014})  have been proposed that could be used to jointly associate the multiple views and to separate the COVID-19 patient groups. For instance,  in \cite{safosida:2021}, we proposed a method that  combined linear discriminant analysis (LDA) and CCA to learn linear representations  that associate the views and separate the classes in each view. However, the relationships among the multiple views and the COVID-19 patient groups are too complex to be understood solely by linear methods.  Nonlinear methods including kernel and deep learning methods could  be used to model  nonlinear structure among the views and between a view and the outcome.

The literature is scarce on nonlinear joint association and separation methods. In \cite{hu2019multi}, a deep neural network method, multi-view linear discriminant analysis network (MvLDAN), was proposed to learn nonlinear projections of multiple views that maximally correlate the views and separate the classes in each view but the convergence of MvLDAN is not guaranteed. Further, MvLDAN and the nonlinear association-based methods for multiple views mentioned above have one major limitation: they do not rank or select features, as such it is difficult to interpret the models and this limits their ability to produce clinically meaningful findings. If we apply MvLDAN or any of  the nonlinear association methods to the COVID-19 omics data, we  will be limited in our ability to identify molecules contributing most to the association of the views and the separation of the COVID-19 patient groups. 

The problem of selecting or ranking features is well-studied in the   statistical learning literature but less-studied in the deep learning literature, especially in deep learning methods for associating multiple views.  In \cite{li2016deep}, a deep feature selection method that adds a sparse one-to-one linear layer between the input layer and the first hidden layer was proposed for feature selection.  In another article \citep{chang2017dropout}, a feature ranking method based on variational dropout was proposed. These methods were developed for data from one view and are not directly applicable to data from multiple views. 
In \cite{TS:2019}, a two-step approach for feature selection using a teacher-student (TS) network was proposed. The ``teacher" step obtains the best low-dimensional representation of the data using  any dimension reduction method (e.g., deep CCA). The ``student" step performs feature selection based on these low-dimensional representations. In particular, a single-layer network with sparse weights is trained to reconstruct the low-dimensional representations obtained from the ``teacher" step, and the features are ranked  based on the weights. This approach is limiting because the model training (i.e., the identification of the low-dimensional representation of the data) and the feature ranking steps are separated as such, one cannot ensure that the top-ranked features identified are  meaningful. 

We propose Deep IDA, (short for Deep Integrative Discriminant Analysis),  a deep learning method for integrative discriminant analysis, to learn complex nonlinear relationships among the multiple molecular data, and between the molecular data and the COVID-19 patient groups. Deep IDA uses deep neural networks (DNN) to nonlinearly transform each view, constructs an optimization problem that takes as input the output from our DNN (i.e., the nonlinearly transformed views), and learns view-specific projections that result in maximum linear correlation of the transformed views and  maximum linear separation within each view. Further, we propose 
a homogeneous ensemble approach for feature ranking where we implement Deep IDA on different training data subsets to yield low-dimensional representations (that are correlated among the views and separate the classes in each view), we aggregate the classification performance from these low-dimensional representations, we rank features based on the aggregates, and we obtain low-dimensional representations of the data based on the top-ranked variables. As a result, Deep IDA permits feature ranking of the views and enhances our ability to identify features from each view that contribute most  to the association of the views and the separation of the classes within each view.  We note that our framework for feature ranking is general and adaptable to many deep learning methods and has potential to yield explainable  deep learning models for associating multiple views. Table \ref{tab:uniquefeatures} highlights the key features of Deep IDA in comparison with some linear and nonlinear methods for multi-view data. Results from our real data application and simulations with small sample sizes suggest that Deep IDA may be a useful method  for small sample size problems compared to other deep learning methods for associating multiple views. 

The rest of the paper is organized as follows. In Section 2, we introduce the proposed method and algorithms for implementing the method. In Section 3, we use simulations to evaluate the proposed method. In Section 4, we use  two real data applications to showcase the performance of the proposed method.  We end with a conclusion remark in Section 5. 

\begin{table} \label{tab:uniquefeatures}
\centering
			\begin{tabular}{lllll}
				\hline
				\hline
				Property/&Linear One-step &  Deep IDA&	Randomized &Deep CCA$^{*}$,\\
 								Methods& Methods& (Proposed) &	KCCA$^{*}$&Deep GCCA$^{+}$\\
		\hline
		\hline
		Nonlinear Relationships & &  \checkmark& \checkmark & \checkmark   \\
		\hline
		Classification& \checkmark& \checkmark&  &    \\
		\hline
		Variable ranking/selection& \checkmark&  \checkmark& &   \\
		\hline
		Covariates&\checkmark & \checkmark & &  \checkmark  \\
		\hline
	One-step&\checkmark & \checkmark&  &  \\
		\hline
		\hline
\end{tabular}
\caption{Unique features of Deep IDA compared to other methods. *Only applicable to two views. +Covariates could be added as additional view in Deep GCCA.}
\end{table}

\section{Method}
Let $\bX^d \in \bR^{n\times p_d}$ be the data matrix for view $d$, $d=1,\ldots,D$ (e.g., proeotimcs, metabolomics, image, clinical data). Each view, $\bX^d$, has $p_d$ variables, all measured on the same set of $n$ individuals or units.  Suppose that each unit belongs to one of two or more classes, $K$. Let $y_i, i=1,\ldots,n$ be the class membership for unit $i$. For each view, let  $\bX^d$ be a concatenation of data from each class, i.e., $\bX^d = [\bX_1^d,\bX_2^d,...,\bX_K^d]^T$, where $\bX_k^d \in \bR^{n_k\times p_d}, k = 1,...,K$ and $n = \sum_{k=1}^K n_k$. For the $k$-th class in the $d$-th view, $\bX_k^d = [\bx_{k,1}^d,\bx_{k,2}^d,...,\bx_{k,n_k}^d]^T$, where $\bx_{k,i}^d \in \bR^{p_d}$ is   a column vector denoting the  view $d$ data values for  the $i$-th unit in the $k$-th class. Given the views and data on class membership, we wish to explore the association among the views and the separation of the classes, and also to predict the class membership of a new unit using the unit's data from all views or  from some of the views. Additionally, we wish to identify the features that contribute most to the overall association among the views and the separation of classes within each view.  Several existing linear (e.g., CCA, generalized CCA) and nonlinear (e.g., deep CCA, deep generalized CCA, random Kernel CCA) methods could be used to first associate the different views to identify low dimensional representations of the views that maximize the correlation among the views or that explain the dependency structure among the views. These low-dimensional representations and the data on class membership could then be used for classification. In this two-step approach, the classification step is independent of the association step and does not take into consideration the effect of  class separation on the dependency structure.  Alternatively, classification algorithms (e.g., linear  or nonlinear methods)  could be implemented on the stacked views, however, this approach ignores the association among the views. Recently,  \cite{safosida:2021} and \cite{zhang2018joint} proposed one-step methods that couple the association step with the separation step and showed that these one-step methods often times result in better classification accuracy when compared to classification on stacked views or the two step methods:  association followed by classification. We briefly review the one-step linear method \citep{safosida:2021} for joint association and classification since it is  relevant to our method.  

\subsection{Integrative Discriminant Analysis (IDA) for joint association and classification}
 \cite{safosida:2021} proposed an integrative discriminant analysis (IDA) method that combines linear discriminant analysis (LDA) and canonical correlation analysis (CCA) to explore linear associations among multiple views and linear separation between classes in each view. Let $\bSb^{d}$ and $\bSw^d$ be the between-class and within-class covariances for the $d$-th view, respectively. That is, $\bSb^{d}=  \frac{1}{n-1} \sum_{k=1}^K n_k (\bmu^d_k - \bmu^d)(\bmu^d_k - \bmu^d)^{\smt}$; $\bSw^d = \frac{1}{n-1}\sum\limits_{k=1}^{K}\sum\limits_{i=1}^{n}(\bx_{ik}^d-{\bmu}_{k})(\bx_{ik}^d-{\bmu}_{k}^d)^{\smt}$, and ${\bmu}_{k}^d = \frac{1}{n_{k}}\sum_{i=1}^{n_{k}}\bx_{ik}^d$ is the mean for class $k$ in view $d$.  Let  $\bS_{dj}, j<d$ be the cross-covariance between the $d$-th and $j$-th views. Let $\mathcal{M}^d=\bS_w^{d^{-1/2}}\bS^d_b\bS_w^{d^{-1/2}}$ and  $\mathcal{N}_{dj}=\bS_w^{d^{-1/2}}\bS_{dj}\bS_w^{j^{-1/2}}$. The IDA method finds  linear discriminant vectors $(\widehat{\bGamma}^1, \ldots, \widehat{\bGamma}^d)$ that maximally associate the multiple views and separate the classes within each view by solving the optimization problem:
\begin{eqnarray} \label{eqn:ida}
	\max_{\bGamma^1,\cdots,\bGamma^D}  \rho\sum_{d=1}^D\text{tr}(\bGamma^{{d\smt}}\mathcal{M}^d \bGamma^d) 
	+ \frac{2(1-\rho)}{D(D-1)}\sum_{d=1,d\ne j }^{D}\text{tr}(\bGamma^{{d\smt}}\mathcal{N}_{dj}\bGamma^j\bGamma^{{j\smt}}\mathcal{N}_{jd}\bGamma^d)
	~\mbox{s.t.}~\text{tr}(\bGamma^{{d\smt}}\bGamma^d)=K-1.
\end{eqnarray}  
The first term in equation  (\ref{eqn:ida}) maximizes the sum of the separation of classes in each view, and the second term maximizes the sum of the pairwise squared correlations between two views. Here, $\rho$ was used to control the influence of separation or association in the optimization problem. The optimum solution for equation (\ref{eqn:ida}) was shown to solve  $D$ systems of eigenvalue problem \citep{safosida:2021}. The discriminant loadings $(\widehat{\bGamma}^1, \ldots, \widehat{\bGamma}^d)$ correspond to the eigenvectors that maximally associate the views and separate the classes within each view.  Once the discriminant loadings were obtained, the views were projected onto these loadings to yield the discriminant scores and samples were classified using nearest centroid. In order to obtain  features contributing most to the association and separation, the authors imposed convex penalties  on $\bGamma^d$ subject to modified eigensystem constraints. In the following sections, we will  modify the IDA optimization problem, cast it as an objective function for deep neural networks to study nonlinear associations among the views and separation of classes within a view. We will also implement a feature ranking approach to identify features contributing most  to the association of the views and the separation of the classes in a view. 


\subsection{Deep Integrative Discriminant Analysis (Deep IDA)}
\begin{figure}[H]
        \centering
        \includegraphics[height=7cm,width=11cm]{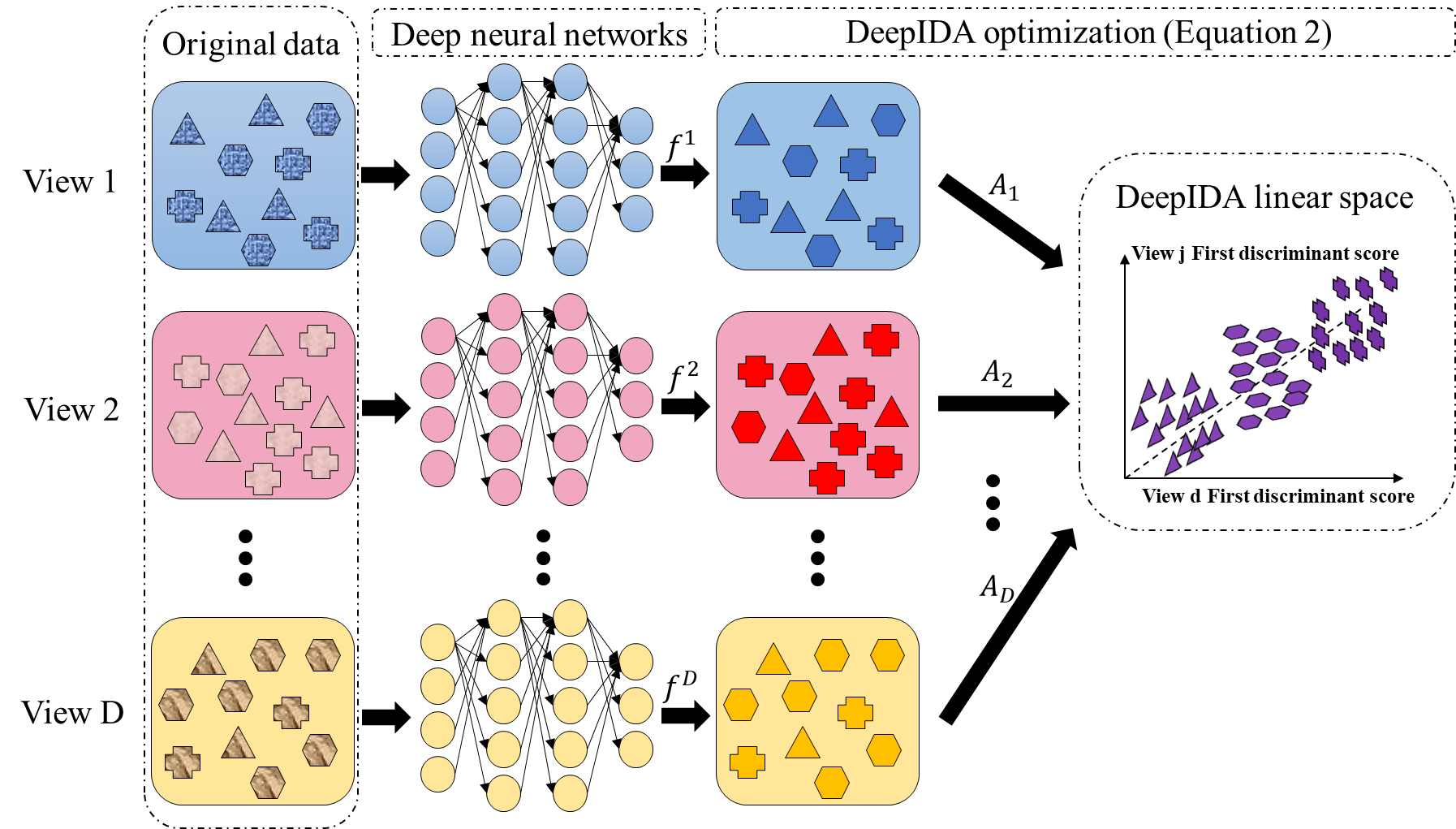}
        \caption{The framework of Deep IDA. Classes are represented by shapes and views are represented by   colors. The deep neural networks (DNN) are used to learn nonlinear transformations of the $D$ views, the outputs of the DNN for the views ($f^d$) are used as inputs in the optimization problem, and we learn linear  projections $\bA_d, d=1,\ldots,D$  that maximally correlate the nonlinearly transformed views and separate the classes within each view.  }
        \label{fig:DeepIDA}
    \end{figure}

We consider a deep learning formulation of the joint association and classification method\citep{safosida:2021} to learn nonlinear relationships among multi-view data and between a view and a binary or multi-class outcome. 
We follow notations in \cite{Andrew:2013} to define our deep learning network. Assume that the deep neural network has $m=1,\ldots,M$  layers for view $d$ (each view can have its own number of layers), and each  layer has $c_m^d$ nodes, for $m=1,\ldots,M-1$. Let $o_1,o_2,...,o_D$ be the dimensions of the final($M$th) layer for the $D$ views. Let $h^d_1 =s(W^d_1 \mathbf{x}^d  + b^d_1) \in \Re^{c^d_1}$ be the output of the first  layer for view $d$. Here, $\mathbf{x}^d$ is a length-$p^d$ vector representing a row in $\bX^d$,  $W^d_1 \in \Re^{ c_1^d \times p^d}$ is a matrix of weights for view $d$, $b^d_1 \in \Re^{c^d_1} $ is a vector of  biases for view $d$ in the first layer,  and  $s \in \Re \longrightarrow \Re$ is a nonlinear activation function. Using $h^d_1$ as an input for the second layer, let the output of the second layer be denoted as $h^d_2 =s(W^d_2h^d_1  + b^d_2) \in \Re^{c^d_2}$, $W^d_2 \in \Re^{c_2^d \times c_1^d}$ and  $b^d_2 \in \Re^{c^d_2}$. Continuing in this fashion, let the output of the $(m-1)$th layer be $h^d_{m-1} =s(W^d_{m-1} h^d_{m-2} + b^d_{m-1}) \in \Re^{c^d_{m-1}}$,  $W^d_{m-1} \in \Re^{c^d_{m-1} \times c^d_{m-2}}$ and  $b^d_{m-1} \in \Re^{c^d_{m-1}}$. Denote the output of the final layer as $f^d(\mathbf{x}^d, \theta^d) = s(W^d_M h^d_{M-1} + b^d_M) \in \Re^{o_d}$, where $\theta^d$ is a collection of all weights, $W^d_m$, and biases, $b^d_m$ for $m=1,\ldots,M$ and $d=1,\ldots,D$. In matrix notation, the output of the final layer of the $d$-th view is denoted as $\bH^d = f^d(\bX^d) \in \Re^{n \times o_d}$, where it is clear that $f^d$ depends on the network parameters. On this final layer, we propose to solve a modified IDA optimization problem to obtain projection matrices that maximally associate the views and separate the classes. Specifically,  we propose to find a set of linear transformations $\bA_d= [\balpha_{d,1},\balpha_{d,2},...,\balpha_{d,l}] \in \bR^{o_d\times l}$, $l\leq \min\{K-1,o_1,...,o_D\}$ such that when the nonlinearly-transformed data are projected onto these linear spaces, the views will have maximum linear association  and the classes within each view will be linearly separated. Figure \ref{fig:DeepIDA} is a visual representation of Deep IDA. 
For a specific view $d$, $\bH^d = [\bH_1^d,\bH_2^d,...,\bH_K^d]^T, \bH_k^d \in \bR^{n_k\times o_d}, k = 1,...,K$ and $n = \sum_{k=1}^K n_k$. For the $k$-th class in the $d$-th final output, $\bH_k^d = [\bh_{k,1}^d,\bh_{k,2}^d,...,\bh_{k,n_k}^d]^T$, where $\bh_{k,i}^d \in \bR^{o_d}$ is a column vector representing  the output for subject $i$ in the $k$th class for view $d$.
Using $\bH^d$ as the data matrix for view $d$, we define the between-class covariance (i.e., $\bS^d_b \in \bR^{o_d\times o_d}$), the total covariance (i.e., $\bS^d_t \in \bR^{o_d\times o_d}$), and the cross-covariance between view $d$ and $j$ ($\bS_{dj}\in \bR^{o_d\times o_j}$ ) respectively as : 
$\bS^d_b = \frac{1}{n-1} \sum_{k=1}^K n_k (\bmu^d_k - \bmu^d)(\bmu^d_k - \bmu^d)^{\smt}$; $\bS^d_t = \frac{1}{n-1} \sum_{i=1}^n  (\bh^d_{k,i} - \bmu^d)(\bh^d_{k,i} - \bmu^d)^{\smt}= \frac{1}{n-1} (\bH^{d^{\smt}}-\bmu^d \cdot \mathbf{1})(\bH^{d^{\smt}}-\bmu^d \cdot \mathbf{1})^{\smt}$, and  $S_{dj} = \frac{1}{n-1} (\bH^{d^{\smt}}-\bmu^d \cdot \mathbf{1})(\bH^{j^{\smt}}-\bmu^j \cdot \mathbf{1})^{\smt}$. Here, $\mathbf{1}$ is an all-ones row vector of dimension $n$, $\bmu_k^d = \frac{1}{n_k}\sum_{i=1}^{n_k} \bh_{k,i}^d \in \bR^{o_d}$ is the $k$-th class mean, and $\bmu^d = \frac{1}{K} \sum_{i=1}^K \bmu_k^d \in \bR^{o_d}$ is the mean for projected view $d$. To obtain the linear transformations $\bA_1,\bA_2,...,\bA_D$ and the parameters of Deep IDA defining the functions $f^d$, (i.e., the weights and biases), we propose to solve the optimization problem: 
\begin{align}\label{eqn:mainopt}
        \argmax_{\bA_1,\ldots,\bA_D, f^1,\ldots,f^D } \left\{ \rho \frac{1}{D} \sum_{d=1}^D tr[\bA_d^T\bS_b^d\bA_d] + (1-\rho) \frac{2}{D(D-1)} \sum_{d=1}^D \sum_{j,j\neq d}^D tr[\bA_d^T \bS_{dj}  \bA_j \bA_j^T \bS_{dj}^T \bA_d] \right\} \nonumber\\
        \mbox{subject to } tr[\bA_d^T\bS_t^d\bA_d] = l, \forall d = 1,...,D,
\end{align}
where $tr[]$ is the trace of some matrix and $\rho$ is a hyper-parameter that controls the relative contribution of the separation and the association to the optimization problem. Here, the first term is an average of the separation for the $D$ views, and the second term is an average of the pairwise squared  correlation between two different views ($\frac{D(D-1)}{2}$ correlation measures in total).

For fixed Deep IDA parameters, (i.e., weights and biases), equation (\ref{eqn:mainopt}) reduces to solving the optimization problem: 
\begin{align}\label{eqn:Asol}
        \argmax_{\bA_1,\ldots,\bA_D} \left\{ \rho \frac{1}{D} \sum_{d=1}^D tr[\bA_d^T\bS_b^d\bA_d] + (1-\rho) \frac{2}{D(D-1)} \sum_{d=1}^D \sum_{j,j\neq d}^D tr[\bA_d^T \bS_{dj}  \bA_j \bA_j^T \bS_{dj}^T \bA_d] \right\} \nonumber\\
        \mbox{subject to } tr[\bA_d^T\bS_t^d\bA_d] = l, \forall d = 1,...,D. 
\end{align}
Denote ${\bS_t^d}^{-\frac{1}{2}}$ as the square root of the inverse of $\bS_t^d$. With the assumption that $o_d < n$, $\bS_t^d$ is non-singular, as such we can take the inverse. Let $\mathcal{M}^d = {\bS_t^d}^{-\frac{1}{2}} \bS_b^d {\bS_t^d}^{-\frac{1}{2}}$, $\mathcal{N}_{dj} = {\bS_t^d}^{-\frac{1}{2}} \bS_{dj} {\bS_t^j}^{-\frac{1}{2}}$ and $\bGamma_d = {\bS_t^d}^{\frac{1}{2}} \bA_d$. Then, the  optimization problem in equation (\ref{eqn:Asol}) is equivalently
\begin{align}\label{eqn:Asol2}
        \argmax_{\bGamma_1,\bGamma_2,...,\bGamma_D} \left\{ \rho \frac{1}{D} \sum_{d=1}^D tr[\bGamma_d^T\mathcal{M}^d\bGamma_d] + (1-\rho) \frac{2}{D(D-1)} \sum_{d=1}^D \sum_{j,j\neq d}^D tr[\bGamma_d^T \mathcal{N}_{dj}  \bGamma_j \bGamma_j^{\smt} \mathcal{N}_{dj}^{\smt} \bGamma_d] \right\} \nonumber\\
        \mbox{subject to } tr[\bGamma_d^{\smt}\bGamma_d] = l, \forall d = 1,...,D,    
\end{align}
and the solution reduces to solving a system of eigenvalue problems. More formally, we have the following theorem. 
\begin{thm}\label{thm:GEVPmain}
	Let $\bS_{t}^{d}$ and $\bS_{b}^{d}$ respectively be the total covariance and the between-class covariance for the top-level representations  $\bH^d, d=1,\ldots,D$. Let $\bS_{dj}$ be the cross-covariance between top-level representations $d$ and $j$.   Assume $\bS_{t}^{d} \succ 0$.  Define $\mathcal{M}^d = {\bS_t^d}^{-\frac{1}{2}} \bS_b^d {\bS_t^d}^{-\frac{1}{2}}$ and  $\mathcal{N}_{dj} = {\bS_t^d}^{-\frac{1}{2}} \bS_{dj} {\bS_t^j}^{-\frac{1}{2}}$. 
	Then $\bGamma^d \in \Re^{o_d \times l}$, $l \le \min\{K-1, o_1,\ldots,o_D\}$ in equation (\ref{eqn:Asol2}) are eigenvectors corresponding respectively to eigenvalues $\bLambda_{d}=$diag$(\lambda_{d_{k}},\ldots,\lambda_{d_{l}})$, $\lambda_{d_{k}} > \cdots > \lambda_{d_{l}}>0$ that iteratively solve the 
 eigensystem problems:
\begin{align*}
   \left(c_1\mathcal{M}^d + c_2\sum_{j\neq d}^D \mathcal{N}_{dj}\bGamma_j\bGamma_j^{\smt}\mathcal{N}_{dj}^{\smt}\right) \bGamma_d &= \bLambda_d \bGamma_d, \forall d = 1,...,D
\end{align*}
where $c_1 = \frac{\rho}{D}$ and $c_2 = \frac{2(1-\rho)}{D(D-1)}$. 
\end{thm}
The proof of Theorem 1 is in the supplementary material. We can initialize the algorithm using any arbitrary normalized nonzero matrix. After iteratively solving $D$ eigensystem problems until convergence, we obtain the optimized linear transformations  $\widetilde{\bGamma}_1,...,\widetilde{\bGamma}_D$ which maximize both separation of classes in the top-level representations, $\bH^d$,   and the association among the top-level representations. Since we find the eigenvector-eigenvalue pairs of  $(c_1\mathcal{M}^d + c_2\sum_{j=1,j\neq d}^D \mathcal{N}_{dj}\bGamma_j\bGamma_j^T\bN_{dj}^T)$, the columns of $\widetilde{\bGamma}_d$, $d = 1,...,D$  are orthogonal and provide unique information that contributes to the association and separation in the top-level representations. Given the optimized linear transformations  $\widetilde{\bGamma}_1,...,\widetilde{\bGamma}_D$, we construct the objective function for the $D$ deep neural networks as: 
\begin{align} \label{eqn:lossobj}
   \argmax_{f^1,f^2,...,f^D} c_1 \sum_{d=1}^D tr[\widetilde{\bGamma}_d^{\smt}\mathcal{M}^d\widetilde{\bGamma}_d] + c_2 \sum_{d=1}^D \sum_{j, j\neq d}^D tr[\widetilde{\bGamma}_d^{\smt} \mathcal{N}_{dj}  \widetilde{\bGamma}_j \widetilde{\bGamma}_j^{\smt} \mathcal{N}_{dj}^T \widetilde{\bGamma}_d].
\end{align}

\begin{thm}\label{thm:loss}
For $d$ fixed, let $\eta_{d,1}, \ldots, \eta_{d,l}$, $l \le \min\{K-1, o_1,\dots,o_D\}$ be the largest $l$ eigenvalues of $c_1\mathcal{M}^d + c_2\sum_{j\neq d}^D \mathcal{N}_{dj}\bGamma_j\bGamma_j^{\smt}\mathcal{N}_{dj}^{\smt}$. Then the solution $\widetilde{f}^d$ to the optimization problem in equation (\ref{eqn:lossobj}) for view $d$ maximizes
\begin{align}
    \sum_{r=1}^l \eta_{d,r}.
\end{align}
\end{thm}
The objective function in Theorem 2 aims to maximize the sum of the $l$ largest  eigenvalues for each view. In obtaining the view-specific eigenvalues, we use the cross-covariances between that view and each of the other views, and the total  and  between-class covariances for that view. Thus, by maximizing the sum of the eigenvalues, we are estimating corresponding eigenvectors that maximize both the association of the views and the separation of the classes within each view. 
By Theorem 2, the solution $\widetilde{f}^1,\ldots,\widetilde{f}^D$, i.e. weights and biases for the $D$ neural networks of the optimization problem (\ref{eqn:lossobj}) is also given by the following:
\begin{align}\label{eqn:lossobj2}
    \argmax_{f^1,f^2,...,f^D} \sum_{d=1}^D \sum_{r=1}^l \eta_{d,r}.
\end{align}

The objectives  (\ref{eqn:Asol2}) and (\ref{eqn:lossobj2}) are naturally bounded because the characteristic roots of every ${\bS^d_t}^{-1}\bS^d_b$ (and hence  ${\bS_t^d}^{-\frac{1}{2}} \bS_b^d {\bS_t^d}^{-\frac{1}{2}}$) is bounded and every squared correlation is also bounded. This guarantees convergent solutions of the loss function in equation (\ref{eqn:lossobj2}) compared to the method in  \citep{dorfer2015deep} that constrain the  within-group covariance and has unbounded loss function. We optimize the objective in  (\ref{eqn:lossobj2}) with respect to the weights and biases for each layer and each view to obtain $\widetilde{f^1},...,\widetilde{f^D}$. The estimates  $\widetilde{f^1(\bX^1)},...,\widetilde{f^D(\bX^D)}$ are used as the low-dimensional representation for classification. For classification, we follow the approach in \cite{safosida:2021} and we use nearest centroid to assign future events to the class with the closest mean. For this purpose, we have the option to  use the pooled low-dimensional representations $\widehat{f}=(\widetilde{f^1(\bX^1)},...,\widetilde{f^D(\bX^D)})$ or the individual estimates $\widetilde{f^d(\bX^d)}, d=1,\ldots,D$. 

\subsection{Optimization and Algorithm}

\begin{itemize}
    \item Feed forward and calculate the loss. The output for $D$ deep neural networks are $\bH^1,...,\bH^D$, which includes the neural network parameters (i.e., the weights and biases). Based on the objective in equation   (\ref{eqn:lossobj2}), the final loss is calculated and denoted as $\mathcal{L} = -\sum_{d=1}^D \sum_{r=1}^l \eta_{d,r}$.

    
    \item Gradient of the loss function.  The loss function $\mathcal{L}$ depends on the estimated linear projections $\widetilde{\bGamma}^d, d=1,\cdots,D$ and since these linear projections use the outputs of the last layer of the network, there are no parameters involved. Therefore we calculate gradient of the loss function with respect to the view-specific output, i.e.,  $\frac{\partial \mathcal{L}}{\partial \bH^d}, d =1,\ldots,D$.
    
    \item Gradient within each sub-network. Since each view-specific sub-network is propagated separately, we can calculate the gradient of each sub-network independently. As the neural network parameters (i.e., weights and biases) of view $d$ network is denoted as $\theta^d$, we can calculate the partial derivative of last layer with respect to sub-network parameters as $\frac{\partial \bH^d}{\partial \theta^d}$. These networks include shallow or multiple layers and nonlinear activation functions, such as Leaky-ReLu \citep{LeakyRelu:2013}.
    \item Deep IDA update via gradient descent. By the chain rule, we can calculate $\nabla_{\theta^d} \mathcal{L} = \frac{\partial \mathcal{L}}{\partial \bH^d} \cdot \frac{\partial \bH^d}{\partial \theta^d}$. We use the \textit{autograd} function in the PyTorch \citep{NEURIPS2019_9015} package to compute this gradient. Therefore, for every optimization step, a stochastic gradient descent-based optimizer, such as Adam \citep{kingma2014adam}, is used to update the network parameters.
\end{itemize}
We describe the Deep IDA algorithm in Algorithm 1. We also describe in Algorithm 2 the approach for obtaining the linear projections using the output of the final layer.  

\begin{algorithm}
    \caption{Algorithm of Deep IDA  }
  \begin{algorithmic}[1]
    \INPUT Training data $\bX = \{\bX^1,\bX^2,..., \bX^D\} $ and class labels $\mathbf{y}$; number of nodes of each layer in $D$ neural networks (including dimensions of linear sub-spaces to project onto, $o_1,o_2,...,o_D$); learning rate $\alpha$
    \OUTPUT  Optimized weights and biases for $D$ neural networks and  corresponding estimates ($\widetilde{f^1(\bX^1)},...,\widetilde{f^D(\bX^D)})$ 
    \STATE \textbf{Initialization} Assign random numbers to weights and biases for $D$ neural networks
    \WHILE{loss not converge}
      \STATE Feed forward the network of each view with latest weights and biases to obtain the final layer $\bH^d = f^d(\bX^d) \in \bR^{n\times o_d}, d=1,2,..,D $
      \STATE Apply Algorithm 2 to obtain $\widetilde{\bGamma}_1,...,\widetilde{\bGamma}_D$
      \STATE Compute eigenvalues of $c_1\mathcal{M}^d + c_2\sum_{j,j\neq d}^D \mathcal{N}_{dj}\widetilde{\bGamma}_j\widetilde{\bGamma}_j^{\smt}\mathcal{N}_{dj}^{\smt}$ to obtain the loss function $-\sum_{d=1}^D \sum_{i=1}^l \eta_{d,i}$
      \STATE Compute the gradient of weights and biases for each network by the PyTorch \textit{Autograd} function
      \STATE Update the weights and biases with the specified learning rate $\alpha$
    \ENDWHILE
  \end{algorithmic}
\end{algorithm}

\begin{algorithm}
    \caption{Algorithm for iteratively solving Eigenvalue Problem}
  \begin{algorithmic}[1]
    \INPUT Training data $\bH = \{\bH^1,\bH^2,..., \bH^D\} $ and corresponding class labels $\mathbf{y}$; convergence criteria $\epsilon$
    \OUTPUT Optimized discriminant loadings $\widetilde{\bGamma}_1,...,\widetilde{\bGamma}_D$
    \STATE Compute $\mathcal{M}^d$, $\mathcal{N}_{dj}$ for $d,j=1,2,..,D $
    \WHILE{$\max_{d=1,..,D} \frac{\|\widehat{\bGamma}_{d_{,new}} - \widehat{\bGamma}_{d_{,old}}\|_F^2}{\| \widehat{\bGamma}_{d_{,old}}\|_F^2}  > \epsilon$}
      \STATE \textbf{for} $d=1,..,D$ \textbf{do}: fix $\widehat{\bGamma}_j, \forall j\neq d$, compute $\widehat{\bGamma}_d$ by \textbf{Theorem 1}
    \ENDWHILE
    \STATE Set $\widetilde{\bGamma}_d=\widehat{\bGamma}_d, \forall d=1,...,D$
  \end{algorithmic}
\end{algorithm}

\subsection{Comparison of Deep IDA with Multi-view Linear Discriminant Analysis Network, MvLDAN}
Our proposed method is related to the method in \cite{hu2019multi} since we  find linear projections of nonlinearly transformed views that separate the classes within each view and maximize the correlation among the views. The authors in \cite{hu2019multi} proposed to solve the following optimization problem for linear projection matrices $\bA_1, \cdots,\bA_D$ and neural network parameters (weights and biases):
\begin{equation}\label{mvldna}
    \argmax_{f^1,\cdots,f^D, \bA_1,\cdots,\bA_D} tr\left((\bS_w + \beta\bA^{\smt}\bA)^{-1}(\bS_b + \lambda\bS_c) \right),
\end{equation}
where $\bA =[\bA_1^{\smt} \cdots \bA^{D^{\smt}}]^{\smt}$ is a concatenation of projection matrices from all views,  $\bS_b$ and $\bS_w$ are the between-class and within-class covariances for all views, respectively,  and $\bS_c$ is the cross-covariance matrix for all views. Further, $\lambda$ and $\beta$ are regularization parameters. The authors  then considered to learn the parameters of the deep neural network by maximizing the smallest eigenvalues of the generalized eigenvalue problem arising from equation (\ref{mvldna})  that do not exceed some threshold that is specified in advance. Although we have the same goal as the authors in \cite{hu2019multi}, our optimization formulation in equation  (\ref{eqn:mainopt}), and our loss function are different. We  constrain the total covariance matrix $\bS_t$ instead of $\bS_w$ and as noted above, our loss function is bounded. As such, we do not have convergence issues when training our deep learning parameters. 
A major drawback of MvLDAN (and existing nonlinear association-based methods for multi-view data) is that they cannot be used to identify variables contributing most to the association among the views and/or separation in the classes. In the next section, we propose an approach that bridges this gap.

\subsection{Feature Ranking  via Bootstrap}
\begin{figure}
        \centering
        \includegraphics[height=16cm,width=16cm]{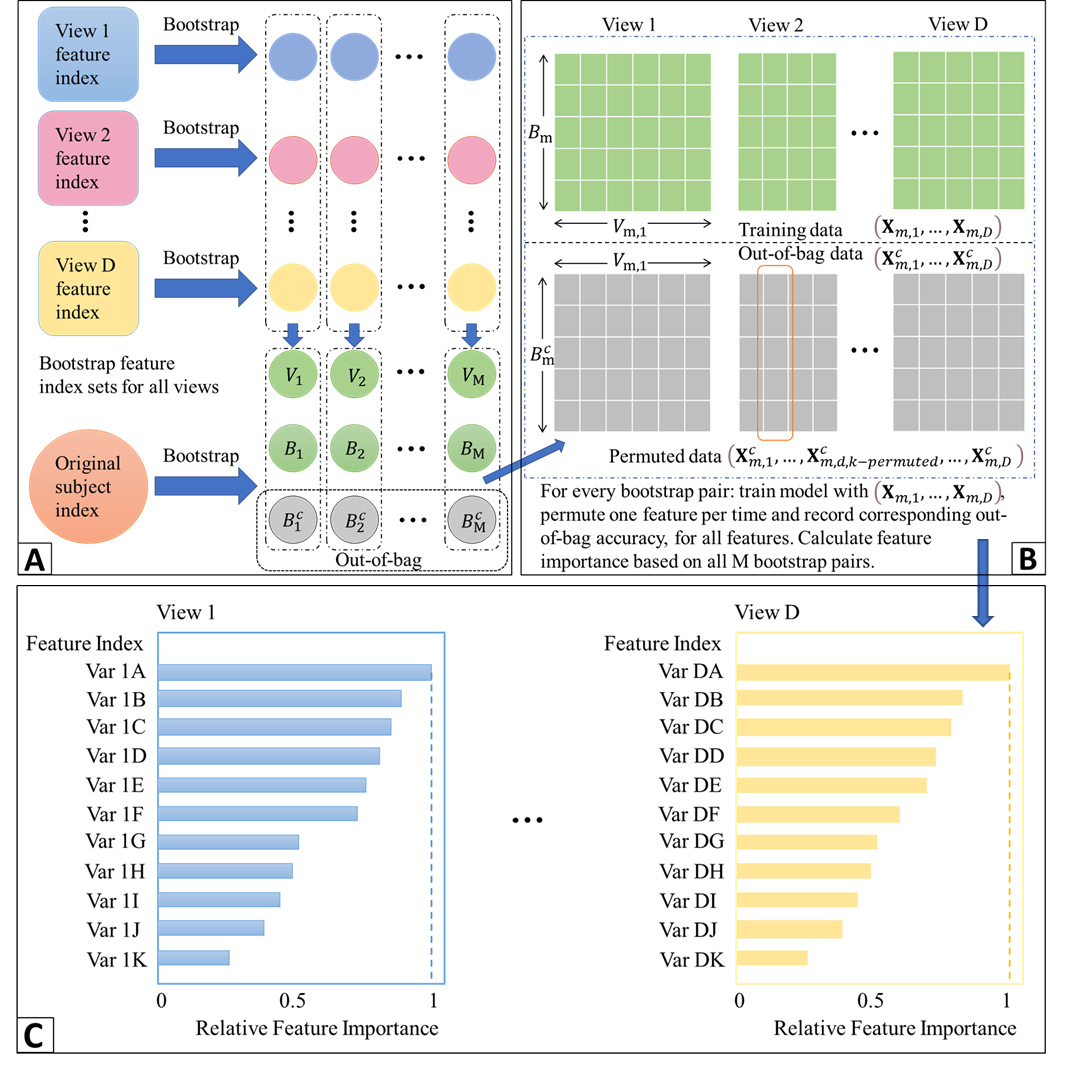}
        \caption{The framework of feature ranking process. A) Bootstrapping samples and features. It includes Steps 1 and 2. $V_m$: the $m$-th bootstrap feature index; $B_m$: the $m$-th bootstrap sample index; $B_m^c$: the $m$-th bootstrap out-of-bag sample index. B) Pairing data, training the reference model, permuting and recording the decrease in classification performance. This includes Steps 3-6. C) Ranking features based on how often the baseline classification accuracy is reduced when permuted. This includes Steps 7 and 8. }
        \label{fig:my_label}
    \end{figure}

A main limitation of existing nonlinear methods for integrating data from multiple views is that it is difficult to interpret the models and this limits their ability to produce clinically meaningful findings. 
We propose a general framework for ranking features in deep learning models for data from multiple views that is based on ensemble learning. Specifically, we propose a homogeneous ensemble approach for feature ranking where we implement Deep IDA on different training data subsets to yield low-dimensional representations (that are correlated among the views while separating the classes in each view), we aggregate the classification performance from these low-dimensional representations, we rank features based on the aggregates, and we obtain a final low-dimensional representations of the data based on the top-ranked variables. We emphasize that while we embed Deep IDA in this feature ranking procedure, in principle, any method for associating multiple views can be embedded in this process. This makes the proposed approach general.  
We outline our feature ranking steps below. Figure \ref{fig:my_label} is a visual representation of the feature ranking procedure.  
\begin{enumerate}
    \item Generate $M$ bootstrap sets of sample indices of the same sample size as the original data by random sampling the indices with replacement. Denote the bootstrap sets of indices as $B_1,B_2,...,B_M$. Let the out-of-bag sets of indices be $B_1^c,B_2^c,...,B_M^c$. In generating the bootstrap training sets of indices, we use stratified random sampling to ensure that the proportions of samples in each class in the bootstrap sets of indices are similar to the original data. 
    \item Draw $q$ bootstrap sets of feature indices for each view. For view $j$, $j=1,\cdots,D$,  draw $0.8p_j$ samples from the index set and denote as $V_{m,j}$. $V_m = \{V_{m,1},V_{m,2},...,V_{m,D}\}$ is the m-th bootstrap feature index for all D views. 
    \item Pair sample  and feature index sets randomly. Denote the pairing results as $(B_1,V_1),(B_2,V_2),...,(B_M,V_M)$. For each pair $(B_m,V_m)$  and $(B_m^c,V_m),(m=1,2,...,M),$ extract corresponding subsets of data. 
    \item  For the $m$th pair, denote the bootstrap data as $\bX_{m,1},...,\bX_{m,D}$ and the out-of-bag data as $\bX_{m,1}^c,...,\bX_{m,D}^c$. Train Deep IDA based on $\bX_{m,1},...,\bX_{m,D}$, and calculate the test classification rate based on $\bX_{m,1}^c,...,\bX_{m,D}^c$. Record this rate as baseline classification rate for pair $m, m=1,2,...,M$.
    \item For the $d$th view in the $m$th pair, permute the $k$th variable in $\bX_{m,d}^c$ and keep all other variables unchanged. Denote the permuted view $d$ data as $\bX_{m,d,k-permuted}^c$. Use the learned model from Step 4 and the permuted data $(\bX_{m,1}^c,...,\bX_{m,d,k-permuted}^c,...,\bX_{m,D}^c)$ to obtain the  classification rate for the permuted data.
    \item Repeat Step 5 for $m=1,...,M$, $d=1,...,D$  and $ k = 1,...,p_d$. Record the  variables that lead to a decrease in classification rate when using the permuted data. 
    \item For the $d$th view, calculate the occurrence proportion of variable $k$, $k=1,2,...,p_d$ (i.e., the proportion of times a variable leads to a decrease in classification accuracy) as $\frac{n_k}{N_k}$, where $n_k$ denotes the number of times that permuting variable $k$ leads to a decrease in out-of-bag classification rate, and $N_k$ denotes the number of times that variable $k$ is permuted (i.e. the total number of times variable $k$ is selected in the bootstrap feature index sets). Repeat this process for all views.
    \item For each view, rank the variables based on the occurrence proportions and select the top-ranked variables as the important variables. The top-ranked variables could be the top $r$ variables or top $r\%$ variables. 
\end{enumerate}
Once we have obtained the top-ranked variables for each view (number  of variables need to be  in advance), we train Deep IDA on the original data but with these top-ranked variables. If testing data are available, say $\bX^d_{test}$, we use the learned neural network parameters to construct the output of the top-level representations for each view, i.e., $\bH^d_{test}=\widetilde{f^d(\bX^d_{test}}), d=1,\dots,D$. These are then  used in a nearest neighbor algorithm to predict the test classes.  
Thus, our final low-dimensional representations of the data are based on features from each view that contribute most  to the association of the views and the separation of the classes within each view.  We implement the proposed algorithm as a Python 3.0 package with dependencies on NumPy \citep{oliphant2006guide} and PyTorch \citep{NEURIPS2019_9015} for numerical computations, and Matlib for model visualization.

\section{Simulations}
\subsection{Set-up}
We conduct simulation studies to assess the performance of Deep IDA for varying data dimensions, and as the relationship between the views and within a view become more complex. We demonstrate that  
Deep IDA is capable of i) simultaneous association of data from multiple views and discrimination of sample classes, and ii) identifying signal variables. 

We consider two different simulation Scenarios. In Scenario One, we simulate data to have linear relationships between views and linear decision boundaries between classes. In Scenario Two, we simulate data to have nonlinear relationships between views and nonlinear decision boundaries between classes. There are $K=3$ classes and  $D=2$ and $D=3$ views in Scenario One. In Scenario Two, there are $K=2$ classes and $D=2$ views. In all Scenarios, we generate 20 Monte Carlo training, validation, and testing sets. We evaluate the proposed and existing methods using the following criteria: i) test accuracy, and ii) feature selection. For feature selection, we evaluate the methods ability to select the true signals. In Scenario One, the first 20 variables are important, and in Scenario Two, the Top $10\%$ of variables in view 1 are signals. Since Deep IDA and the teacher-student (TS) framework rank features, and SIDA assigns zero weights to unimportant variables, for fair comparison, we only assess the number of signal variables that are in the Top 20 (for Scenario One) and the Top $10\%$ (for Scenario Two) variables selected by the methods. We compare test accuracy for Deep IDA with and without the variable ranking approach proposed in this manuscript. 

\subsection{Comparison Methods}
We compare Deep IDA with classification-, association-, and joint association and classification-based methods. For classification-based methods, we consider the support vector machine \citep{Hastie:2009} on stacked views. For association-based methods, we consider nonlinear methods such as deep canonical correlation analysis (Deep CCA) \citep{Andrew:2013}, deep generalized CCA (DGCCA) \citep{Benton2:2019} when there are three or more views, and randomized kernel CCA (RKCCA) \citep{pmlr-v32-lopez-paz14}. The association-based methods only consider nonlinear associations between views, as such we follow  with SVM to perform classification using the learned low-dimensional representations from the methods. We also compare Deep IDA to SIDA \citep{safosida:2021}, a joint association and classification method that models linear relationships between views and among classes in each view. We perform SIDA and RKCCA using the Matlab codes the authors provide. We set the number of random features in RKCCA as 300 and we select the bandwidth of the radial basis kernel  using median heuristic \citep{garreau2017large}.  We perform Deep CCA and Deep generalized CCA using the PyTorch codes the authors provide. We couple Deep CCA and Deep GCCA with the teacher-student framework (TS) \citep{TS:2019} to rank variables. We also investigate the performance of these methods when we use variables selected from Deep IDA.

\subsection{Linear Simulations}
We consider two simulation settings in this Scenario and we simulate data similar to simulations in \cite{safosida:2021}. In Setting One, there are $D=2$ views $\bX^{1}$ and $\bX^{2}$, with $p_1=1,000$ and $p_2=1,000$ variables respectively. There are $K=3$ classes each with size $n_k=180$, $k=1,2,3$ giving a total sample size of $n=540$.  Let  $\bX^{d} = [\bX_1^d, \bX_2^d, \bX_3^d], d=1,2$ be a concatenation of data from the three classes. The combined data $\left(\bX^{1}_k, \bX^{2}_k\right)$ for each class are simulated from $N(\bmu_k, \bSigma)$, $\bmu_k = (\bmu^1_k, \bmu^2_k)^{\smt} \in \Re^{p_1 + p_2}, k=1,2,3$ is the combined mean vector for class $k$; $\bmu^1_k \in \Re^p_1, \bmu^2_k \in \Re^p_2$ are the mean vectors for $\bX^{1}_k$ and $\bX^{2}_k$ respectively. The covariance matrix  $\bSigma$ for the combined data for each class is partitioned as \begin{eqnarray}
\bSigma =\left(
\begin{array}{cc}
\bSigma^{1} & \bSigma^{12} \nonumber\\
\bSigma^{21} & \bSigma^{2} \nonumber\
\end{array} \right), \bSigma^1 =\left(
\begin{array}{cc}
\widetilde{\bSigma}^{1} & \textbf{0} \nonumber\\
\textbf{0} & \bI_{p_1-20} \nonumber\
\end{array} \right), \bSigma^2 =\left(
\begin{array}{cc}
\widetilde{\bSigma}^{2} & \textbf{0} \nonumber\\
\textbf{0} & \bI_{p_2-20} \nonumber\
\end{array} \right)
\end{eqnarray}

\noindent where $\bSigma^{1}$,  $\bSigma^{2}$ are respectively the covariance of  $\bX^{1}$ and $\bX^{2}$, and $\bSigma^{12}$ is the cross-covariance between the two views. We generate $\widetilde{\bSigma}^{1}$ and $\widetilde{\bSigma}^{2}$ as block diagonal with 2 blocks of size 10, between-block correlation 0, and  each block is a compound symmetric matrix with correlation 0.8. We generate the cross-covariance matrix $\bSigma^{12}$ as follows.  Let $\mathbf{V}^{1} = [\mathbf{V}^{1}_{1},~ \mathbf{0}_{(p_1-20) \times 2}]^{\smt} \in \Re^{p_1 \times 2 }$ and  the entries of $\bV^{1}_{1} \in \Re^{20 \times 2}$ are \textit{i.i.d} samples from U(0.5,1). We similarly define $\mathbf{V}^2$ for the second view,  and we  normalize  such that $\mathbf{V}^{1^{T}}\bSigma^{1}\mathbf{V}^{1} = \mathbf{I}$ and $\mathbf{V}^{2^{\smt}}\bSigma^2 \mathbf{V}^{2} = \mathbf{I}$. We then set  $\bSigma^{12} = \bSigma^{1}\bV^1\bD\bV^{2^{\smt}} \bSigma^{2}$, $\bD= \text{diag}(0.4, 0.2)$ to depict moderate association between the views. For class separation, define the matrix $[\bSigma\bA, \textbf{0}_{(p_1 + p_2)}] \in \Re^{(p_1 + p_2) \times 3}$; $\bA=[\bA^1, \bA^2]^{\smt} \in \Re^{(p_1 +p_2) \times 2}$, and set the first, second, and third columns as the mean vector for class 1, 2, and 3, respectively. 
Here, the first column of $\bA^{1} \in \Re^{p_1 \times 2}$ is set to $(c\textbf{1}_{10}, \textbf{0}_{p_1-10})$ and
the second column is set to $( \textbf{0}_{10},-c\textbf{1}_{10}, \textbf{0}_{p-20})$; we fix $c$ at 0.2. We set $\bA^{2} \in \Re^{p_2 \times 2}$ similarly, but we fix $c$ at $0.1$ to allow for different class separation in each view. 

In Setting Two, we simulate $D=3$ views, $\bX^d, d=1,2,3$, and each view is a concatenation of data from three classes as before. The combined data $\left(\bX^{1}_k, \bX^{2}_k,\bX^{3}_k\right)$ for each class are simulated from $N(\bmu_k, \bSigma)$, where $\bmu_k = (\bmu^1_k, \bmu^2_k,\bmu^3_k)^{\smt} \in \Re^{p_1+p_2+p_3}, k=1,2,3$ is the combined mean vector for class $k$; $\bmu^d_k \in \Re^{p_d}, j=1,2,3$ are the mean vectors for $\bX^{d}_k, d=1,2,3$. The true covariance matrix $\bSigma$ is defined similar to Setting One but with the following modifications. We include $\bSigma_{3}$, $\bSigma_{13}$, and $\bSigma_{23}$, and we set $\bSigma_{13}=\bSigma_{23}=\bSigma_{12}$. Like $\bSigma_{1}$ and $\bSigma_{2}$,  $\bSigma_{3}$ is partitioned into signals and noise, and the covariance for the signal variables, $\widetilde{\bSigma}^{3}$, is  also block diagonal with 2 blocks of size 10, between-block correlation 0, and  each block is  compound symmetric matrix with correlation 0.8. We take $\bmu_k$ to be the columns of  $[\bSigma\bA, \textbf{0}_{(p_1+p_2 +p_3)}] \in \Re^{(p_1+p_2 +p_3) \times 2}$, and $\bA=[\bA^1, \bA^2,\bA^3]^{\smt} \in \Re^{(p_1 +p_2+p_3) \times 2}$. The first column of $\bA^{j} \in \Re^{p_j \times 2}$ is set to $(c_j\textbf{1}_{10}, \textbf{0}_{p_1-10} )$ and the second column is set to  $( \textbf{0}_{10},-c_j\textbf{1}_{10}, \textbf{0}_{p-20})$ for $j=1,2,3$. We set $(c_1,c_2,c_3) =(0.2,0.1,0.05)$ to allow for different class separation in each view.

\subsubsection{Results for Linear Simulations}
Table \ref{tab: Linear} gives classification accuracy for the methods  and the true positive rates for the top 20 variables selected. We implemented a three-hidden layer network with dimensions 512, 256, and 64 for both Deep IDA and Deep CCA. The dimension of the output layer was set as 10.  Table 3 in the supplementary material lists the network structure used for each setting.
For Deep IDA + Bootstrap, the bootstrap algorithm proposed in the Methods Section was implemented on the training data to choose the top 20 ranked variables. We then implemented Deep IDA on the training data but with just the variables ranked in the top 20 in each view. The learned model and the testing data were used to obtain test error.  To compare our feature ranking process with the teacher-student (TS) network approach for feature ranking, we also implemented Deep IDA without the bootstrap approach for feature ranking, and we used the learned model from Deep IDA in the TS framework for feature ranking.  We also performed feature ranking using the learned model from Deep CCA (Setting One) and Deep GCCA (Setting Two). The average error rates for the nonlinear methods are higher than the error rate for SIDA, a  linear method for joint association and classification analysis. This is not surprising as the true relationships among the views, and the classes within a view are linear. Nevertheless, the average test error rate for Deep IDA that is based on the top 20 variables in each view from the bootstrap method (i.e., Deep IDA + Bootstrap) is lower than the average test error rates from Deep CCA, RKCCA, and SVM (on stacked views). When we implemented Deep CCA, RKCCA, SVM, and DGCCA on the top 20 variables that were selected by our proposed method, we observed a decrease in the error rate across most of the methods, except for RKCCA. For instance, the error rates for Deep CCA using all variables compared to using the top 20 variables identified by our method were $33.17\%$ and $22.95\%$, respectively.  Further, compared to Deep IDA on all variables (i.e., Deep IDA + TS), Deep IDA + Bootstrap has a lower average test error, demonstrating the advantage of variable selection. In  Setting Two, the classification accuracy for Deep GCCA was poor.  In terms of variable selection, compared to SIDA, the proposed method was competitive at identifying the top-ranked 20 variables. The TS framework for ranking variables was sub-optimal as evident from the true positive rates for  Deep IDA + TS, Deep CCA + TS, and Deep GCCA + TS. 



\begin{table}
\spacingset{1}
\begin{small}
\begin{centering}

				\caption{Linear Setting:  RS; randomly select tuning parameters space to search.  TPR-1; true positive rate for $\mathbf{X}^1$. Similar for TPR-2. TS: Teacher student network. $-$ indicates not applicable. Deep IDA + Bootstrap is when we use the bootstrap algorithm to choose the top 20 ranked variables, train Deep IDA with the top 20-ranked variables, and then use the learned model and the test data to obtain test errors.}	\label{tab: Linear}
				\begin{tabular}{lrrrr}
			\hline
			\hline
			Method&Error (\%)& 	TPR-1&TPR-2 &TPR-3	\\
			\hline
			\hline
			\textbf{Setting One}&  \\
			\rb{Deep IDA + Bootstrap}	&24.69 & 	100.00&95.25& -\\
			{Deep IDA+ TS}	&33.87& 	33.25&21.75& -\\
				{SIDA}	&20.81& 	99.50&93.50& -\\
			Deep CCA + TS & 33.17&	4.25&3.25& -\\
			Deep CCA on selected top 20 variables & 22.95 &	-&-& -\\
			RKCCA&40.1& 	-&-& -\\
			RKCCA on selected top 20 variables &42.07& 	-&-& -\\
			SVM (Stacked views) & 31.53&	-&-& -\\
			SVM on selected top 20 variables (Stacked views)  & 22.03&	-&-& -\\

			\hline				
			\textbf{Setting Two}&  \\
		\rb{Deep IDA + Bootstrap}	&23.16 &100.00&94.75& 78.75\\
				{Deep IDA + TS}	&31.22& 	72.00&57.75& 47.75\\
				{SIDA}	&19.79& 	99.75&99.50& 97.25\\
			DGCCA + TS &60.01 &	2.0&2.0& 2.25\\
			DGCCA on selected top 20 variables & 57.40&	-&-& -\\
			SVM (Stacked views) &29.06&	-&-& -\\
						SVM on selected top 20 variables (Stacked views) & 19.56 &	-&-& -\\

			\hline
			\hline
		\end{tabular}
		\end{centering}
		\end{small}
\end{table}	

\subsection{Nonlinear Simulations}
	\begin{small}
		\begin{figure}[H]
			\begin{center}
				\begin{tabular}{ll}
					\includegraphics[height = 1.5in]{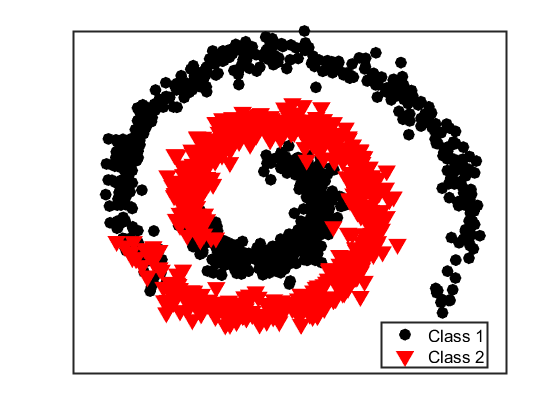}& \includegraphics[height = 1.5in]{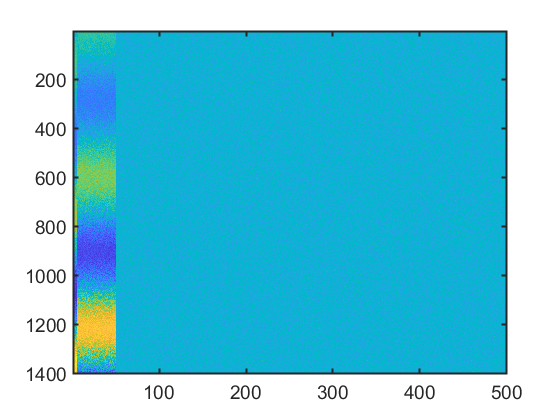}\\
				\end{tabular}
				 \caption{Setting One. (Left panel) Structure of nonlinear relationships between signal variables in view 1. (Right panel) Image plot of view 1 showing the first 50 variables as signals.  }
    \label{fig:nonlinears1}
			\end{center}
		\end{figure}
	\end{small}
We consider four different settings for this scenario. Each setting has $K=2$ classes but they vary in their dimensions. In each setting, $10\%$ of the variables in the first view are signals and the first five signal variables in the first view are related to the remaining signal variables in a nonlinear way (See Figure \ref{fig:nonlinears1}). We generate data for View 1 as: 
$\mathbf{X}_1= \widetilde{\bX}_1 \cdot \bW + 0.2\bf{E}_1$ where $(\cdot)$ is element-wise multiplication,  $\bW \in \Re^{n \times p_1}= [\mathbf{1}_{0.1\times p_1}, \mathbf{0}_{0.9\times p_1}]$ is a matrix of ones and zeros, $\mathbf{1}$ is a matrix of ones, $\mathbf{0}$ is matrix of zeros, and $\mathbf{E}_1 \sim N(0,1)$.  The first five columns (or variables) of $\widetilde{\bX}_1 \in \Re^{n \times p^1}$ are simulated from $\exp(0.15\btheta)\cdot\sin(1.5\btheta)$. The next $0.1p^1 - 5$ variables are simulated from $\exp(0.15\btheta)\cdot\cos(1.5\btheta)$. Here, $\btheta=\tilde{\btheta} + 0.5U(0,1)$, and $\tilde{\btheta}$ is a vector of $n$ evenly spaced points between $0$ and $3\pi$. The remaining $0.9p^1$ variables (or columns) in  $\widetilde{\bX}_1$ are generated from the standard normal distribution. View 2 has no signal variables and the variables do not have nonlinear relationships.  Data for View 2 are generated as follows. We set each variable in $\mathbf{X}_1$ with negative entries to zero, normalized each variable to have unit norm and added a random number generated from the standard uniform distribution.

\subsubsection{Results for Nonlinear Simulations}
Table \ref{tab: nonlinear2} gives the classification and variable selection accuracy.
We chose the number of layers that gave the minimum validation error (based on our approach without bootstrap) or better variable selection. 
Table 4 in the supplementary material lists the network structure used for each setting. 
We compare Deep IDA to the nonlinear methods. 
Similar to the linear setting, for Deep IDA + Bootstrap, we implemented the bootstrap approach for variable ranking on the training data to choose the top $10\%$ ranked variables. We then implemented Deep IDA on the training data but with just the selected variables. The learned model and the testing data were used to obtain test classification accuracy. We also implemented Deep CCA, RKCCA, and SVM with the variables that were selected by Deep IDA + Bootstrap to facilitate comparisons.  For Deep IDA + TS and Deep CCA + TS, we implemented the teacher-student algorithm to rank the variables. Since in this Scenario, only view 1 had informative features, we expected the  classification accuracy  from view 1 to be better than the  classification accuracy from both views and this is what we observed across most methods. We  note that when training the models, we used both views. The classification accuracy from Deep IDA was generally higher than the other methods, except in Setting Three where it was lower than Deep CCA on the whole data (i.e., Deep CCA + TS). We compared the classification accuracy of the proposed method with  (i.e., Deep IDA + Bootstrap) and without feature ranking by our method (i.e., Deep IDA + TS) to assess the effect variable selection has on classification estimates from our deep learning models. Deep IDA + Bootstrap had competitive or better classification accuracy (especially when using view 1 only for classification)  compared to Deep IDA + TS.  Further, the classification accuracy  for Deep IDA + Bootstrap was generally higher than the other methods applied to data with variables selected by Deep IDA + Bootstrap (e.g., Deep CCA on top 50 selected features, Setting One).   
SVM applied on both views stacked together and on just view 1, either using the whole data or using data with variables selected by  Deep IDA,  resulted in similar classification performance, albeit lower than the proposed method. Thus, in this example, although only view 1 had signal variables, the classification performance from using both views was better than using only view 1 (e.g., SVM on view 1), attesting to the benefit of multi-view analyses. 
In terms of  variable selection, the TS framework applied on Deep IDA and Deep CCA yielded sub-optimal results. 

Taken together,  the classification and variable selection accuracy from both the linear and nonlinear simulations suggest that the proposed method is capable of ranking the signal variables higher, and is also able to yield competitive or better classification performance, even in situations where the sample size is less than the number of variables. 
\begin{table}
\spacingset{1}
\begin{centering}
	\begin{footnotesize}	
				\caption{Mean (std.error) accuracy and true positive rates. View 1 data have signal variables with  nonlinear relationships. TPR-1; true positive rate for $\mathbf{X}^1$. Deep IDA/Deep CCA/RKCCA view 1 means using the discriminant scores of view 1 only for classification. SVM view 1 uses view 1 data to train and test the model. $-$ indicates not applicable}	\label{tab: nonlinear2}
				\begin{tabular}{lll}
			\hline
			\hline
			Method& Mean (\%) Accuracy& 	TPR-1	\\
			\hline
			\hline
			\textbf{Setting One}&  \\
	$(p_1=500,p_2=500, n_1=200, n_2=150)$\\
		\rb{Deep IDA + Bootstrap }	&60.87 (1.28) & 100.0 \\
		\rb{Deep IDA + Bootstrap } view 1	&81.17 (2.89) & 100.0 \\
		{Deep IDA + TS}	&62.60 (2.02) & 10.20	 \\
		 {Deep IDA + TS View 1}	&81.49 (3.06)  &10.20\\
			Deep CCA + TS & 58.20(0.59)&	8.30\\
			Deep CCA + TS view 1 & 61.26(0.77)&	8.30\\
			Deep CCA on top 50 selected features & 58.91(0.82)&	-\\
			Deep CCA on top 50 selected features view 1 & 59.87(0.87)&	-\\
			RKCCA&61.06 (0.47)&	-\\
			RKCCA View 1&64.93 (0.63)&	-\\
			RKCCA on top 50 selected features &56.21 (0.77)&	-\\
			RKCCA View 1 on top 50 selected features &58.94 (0.78)&	-\\
			SVM &54.20(0.30)&	-\\
		    SVM view 1 &50.26(0.21)&	-\\
		    SVM on top 50 selected features & 50.37(0.27)&	-\\
			SVM on top 50 selected features view 1 & 50.07(0.26)&	-\\
			\hline			
		 \textbf{Setting Two}&  \\
		$(p_1=500,p_2=500, n_1=3,000, n_2=2,250)$\\
		\rb{Deep IDA + Bootstrap }	&89.45(2.16)& 63.70\\
		\rb{Deep IDA + Bootstrap  View 1}	&90.49(2.25)  & 63.70 \\
		{Deep IDA + TS}	&91.78 (1.73) & 10.50	 \\
		 {Deep IDA + TS} View 1	&84.37 (1.49) &10.50\\
			Deep CCA + TS & 59.84(0.40)&	33.70\\
			Deep CCA + TS view 1 & 60.35(0.34)&	33.70\\
			Deep CCA on top 50 selected features & 58.50(0.52)&	-\\
			Deep CCA on top 50 selected features view 1 & 58.32(0.52)&	-\\
 			RKCCA&57.14 (0.00)&	-\\
			RKCCA View 1&57.14 (0.00)&	-\\
			RKCCA on top 50 selected features &66.30 (0.96)&	-\\
			RKCCA View 1 on top 50 selected features &66.32 (0.98)&	-\\
			SVM &54.42(0.11)&	-\\
		    SVM view 1 &52.81(0.13)	\\
		    SVM on top 50 selected features & 50.56(0.07)&	-\\
			SVM on top 50 selected features view 1 & 50.49(0.04)&	-\\
			\hline
				\textbf{Setting Three}&  \\
		$(p_1=2000,p_2=2000, n_1=200, n_2=150)$\\
		\rb{Deep IDA + Bootstrap }	&54.77(0.91) & 96.05\\
		\rb{Deep IDA + Bootstrap  View 1}	&70.40(2.17) & 96.05 \\
		{Deep IDA + TS}	&61.67 (1.74) & 	30.55 \\
		 {Deep IDA + TS} View 1	&60.73 (1.76) &30.55\\
			Deep CCA + TS & 62.27(0.46)&	10.28\\
			Deep CCA + TS view 1 & 70.83(0.36)&	10.28\\
			Deep CCA on top 200 selected features & 61.43(0.62)&	-\\
			Deep CCA on top 200 selected features view 1 & 68.67(0.83)&	-\\
			RKCCA&60.24 (0.63)&	-\\
			RKCCA View 1&63.54 (0.57)&	-\\
			RKCCA on top 200 selected features &58.70 (0.52)&	-\\
			RKCCA View 1 on top 200 selected features &61.90 (0.91)&	-\\
			SVM &54.69(0.45)&	-\\
		    SVM view 1 &53.97(0.41)&	-\\
		    SVM on top 200 selected features & 51.14(0.42)&	-\\
			SVM on top 200 selected features view 1 & 50.19(0.35)&	-\\
		    \hline
				\textbf{Setting Four}&  \\
		$(p_1=2000,p_2=2000, n_1=3,000, n_2=2,250)$\\
		\rb{Deep IDA + Bootstrap }	&69.38(1.44)& 83.20\\
		\rb{Deep IDA + Bootstrap  View 1}	&71.34(1.66)  & 83.20 \\
		{Deep IDA + TS}	&64.52 (1.32) & 10.78	 \\
		 {Deep IDA + TS} View 1	&69.99 (2.88) &10.78\\
		Deep CCA + TS & 60.33(0.76)&	33.48\\
			Deep CCA + TS view 1 & 58.54(1.00)&	33.48\\	
			Deep CCA on top 200 selected features & 59.58(1.66)&	-\\
			Deep CCA on top 200 selected features view 1 & 59.28(1.71)&	-\\
			RKCCA&57.14 (0.00)&	-\\
			RKCCA View 1&57.14 (0.00)&	-\\
			RKCCA on top 200 selected features &55.30 (0.79)&	-\\
			RKCCA View 1 on top 200 selected features &61.21 (0.81)&	-\\
			SVM &53.13(0.16)&	-\\
		    SVM view 1 &53.37(0.13)&	-\\
		    SVM on top 200 selected features & 52.72(0.13)&	-\\
			SVM on top 200 selected features view 1 & 50.59(0.12)&	-\\
			\hline		
			\hline
		\end{tabular}
	\end{footnotesize}
\end{centering}	
\end{table}

\section{Real Data Analyses}
We consider two real datasets: a) handwriting image data, and b)  COVID-19 omics data. The image data will be used to primarily assess the classification performance of the proposed method without feature ranking while the COVID-19  data will be used to assess classification performance and to also demonstrate that Deep IDA is capable of identifying biologically relevant features. 

\subsection{Evaluation of the Noisy MNIST digits data}
\begin{figure}[H]
\centering
\begin{tabular}{cc}
         \includegraphics[width=0.3\textwidth]{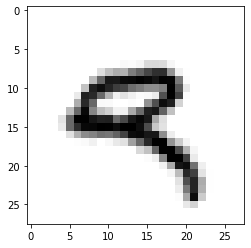}&
         \includegraphics[width=0.3\textwidth]{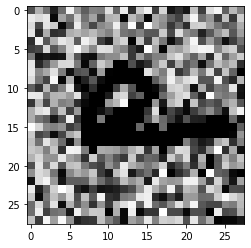}\\
\end{tabular}
    \caption{An example of Noisy MNIST data. For the subject with label "9", view 1 observation is on the left and view 2 observation is on the right.}
    \label{fig:mnist}
\end{figure}

The original MNIST handwritten image dataset \citep{MNIST:1998} consists of 70,000 grayscale images of handwritten digits split into training, validation and testing sets of 50,000, 10,000 and 10,000 images, respectively. The validation set was used to select network parameters from the best epoch (lowest validation loss). Each image is $28 \times 28$ pixels and has associated with it a label that denotes which digit the image represents (0-9).  In \cite{Wang:2015}, a more complex and challenging noisy version of the original data was generated and used as a second view. First,  all pixel values were scaled to 0 and 1.  The images were randomly rotated at angles uniformly sampled from $[-\frac{\pi}{4},\frac{\pi}{4}]$, and the resulting images were used as view 1.  Each rotated image  was paired with an image of the same label from the original MNIST data, independent random noise generated from U(0,1) was added, and the pixel values were truncated to [0,1]. The transformed data is view 2. Figure  \ref{fig:mnist} shows two image plots of a digit for views 1 and 2.  Of note,  view 1 is informative  for classification and view 2 is noisy. Therefore, an ideal multi-view classification method should be able to extract the useful information from view 1 while disregarding the noise in view 2. 

The goal of this application is to evaluate how well the proposed method without feature ranking can classify the digits using the two views. Thus,  we applied Deep IDA without feature ranking  and the competing methods to the training data and  we used the learned models and the testing data to obtain test classification accuracy. The validation data was used in Deep IDA and Deep CCA to choose the best model among all epochs. Table 5 in the supplementary material lists the network structure used in this analysis.  Table \ref{tab: mnist} gives the test classification results of the methods. We observe that the test classification accuracy of the  proposed method with nearest centroid classification is better than SVM on the stacked data,  and is comparable to Deep CCA. We observe a slight advantage of the proposed method when we implement SVM on the final layer of Deep IDA. 

\begin{table}
\spacingset{1}
\centering
				\caption{Noisy MNIST data: SVM was implemented on the stacked data. For Deep CCA + SVM, we trained  SVM on the combined outputs (from view 1 and view 2) of the last layer of Deep CCA. For Deep IDA + NCC, we implemented the Nearest Centroid Classification on the combined outputs (from view 1 and view 2) of the last layer of Deep IDA. For Deep IDA + SVM,  we trained  SVM on the combined outputs (from view 1 and view 2) of the last layer of Deep IDA.	\label{tab: mnist}}
				\begin{tabular}{lr}
			\hline
			\hline
			Method&Accuracy (\%)  	\\
			\hline
			\hline
            SVM (combined view 1 and 2) &93.75\\
			Deep CCA + SVM & 97.01\\
			\rb{Deep IDA + NCC} & 97.74\\
			\rb{Deep IDA + SVM}& 99.15\\
			RKCCA + SVM &  91.79 \\
			\hline
			\hline
		\end{tabular}
\end{table}

\subsection{Evaluation of the COVID-19 Omics Data}
\subsubsection{Study Design and Goal}
The molecular and clinical data we used are described in  \cite{overmyer2021large}. Briefly, blood samples from 128 patients admitted to the Albany Medical Center, NY from 6 April 2020 to 1 May 2020 for moderate to severe respiratory issues collected. These samples were quantified for metabolomics, RNA-seq, proteomics, and lipidomics data.  In addition to the molecular data, various demographic and clinical data were obtained at the time of enrollment. For eligibility, subjects had to be at least 18 years and admitted to the hospital for COVID-19-like symptoms. Of those eligible, $102$ had COVID-19, and $26$ were without COVID-19. 
Of those with COVID-19, 51 were admitted to the Intensive Care Unit (ICU) and 51 were not admitted to the ICU (i.e., Non-ICU). Of those without COVID-19, 10 were Non-ICU patients and 16 were ICU patients. 
Our goal is to elucidate the molecular architecture of COVID-19 severity by identifying molecular signatures that are associated with each other and have  potential to  discriminate  patients with and without COVID-19 who were or were not admitted to the  ICU. 

\subsubsection{Data pre-processing and application of Deep IDA and competing methods}
Of the 128 patients, 125 had both omics and clinical data. We focused on proteomics, RNA-seq, and metabolomics data in our analyses since many lipids were not annotated.  We formed a four-class classification problem using COVID-19 and ICU status. Our four groups were: with COVID-19 and not admitted to the ICU (COVID Non-ICU), with COVID-19 and admitted to the ICU (COVID ICU), no COVID-19 and admited to the ICU (Non-COVID ICU), and no COVID-19 and not admitted to the ICU (Non-COVID Non-ICU). The frequency distribution of samples in these four groups were: $40\%$ COVID ICU, $40\%$ COVID Non-ICU, $8\%$ Non-COVID Non-ICU, and  $12\%$ Non-COVID ICU. The initial dataset contained $18,212$ genes, $517$ proteins, and $111$ metabolomics features. Prior to applying our method, we pre-processed the data (see Supplementary Material) to obtain a final dataset of $\mathbf{X}^{1} \in \Re^{125 \times 2,734}$ for the RNA-sequencing data, $\mathbf{X}^{2} \in \Re^{125 \times 269}$ for the protoemics data, and $\mathbf{X}^{3} \in \Re^{125 \times 66}$ for the metabolomics. 

We randomly split the data into training ($n=74$) and testing ($n=51$) sets while keeping the proportions in each group similar to the original data, we applied the methods on the training data and we assessed error rate using the test data. We evaluated Deep IDA with and without feature selection. For Deep IDA with feature selection, we obtained the top 50 and top 10\% molecules after implementing Algorithm 1, learned the model on the training data with only the molecules that were selected, and  estimated the test error with the testing data. We also assessed the performance of the other methods using variables that were selected by Deep IDA. This allowed us to investigate the importance of feature selection for these methods. 

\subsubsection{Test Accuracy and Molecules Selected}
\begin{figure}[H]
	\centering
	\begin{tabular}{cc}
		\includegraphics[width=0.70\textwidth]{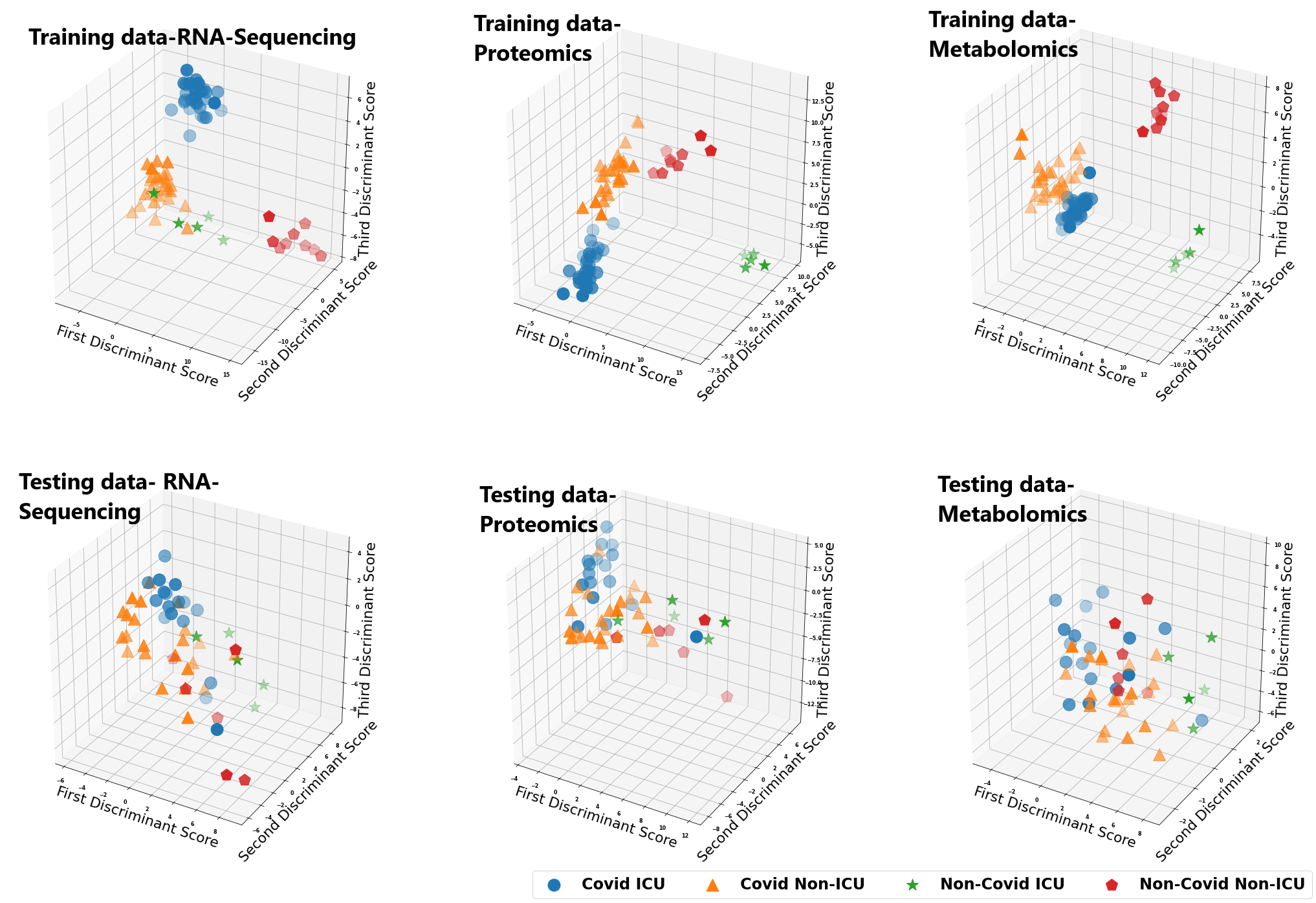}\\
		\includegraphics[width=0.70\textwidth]{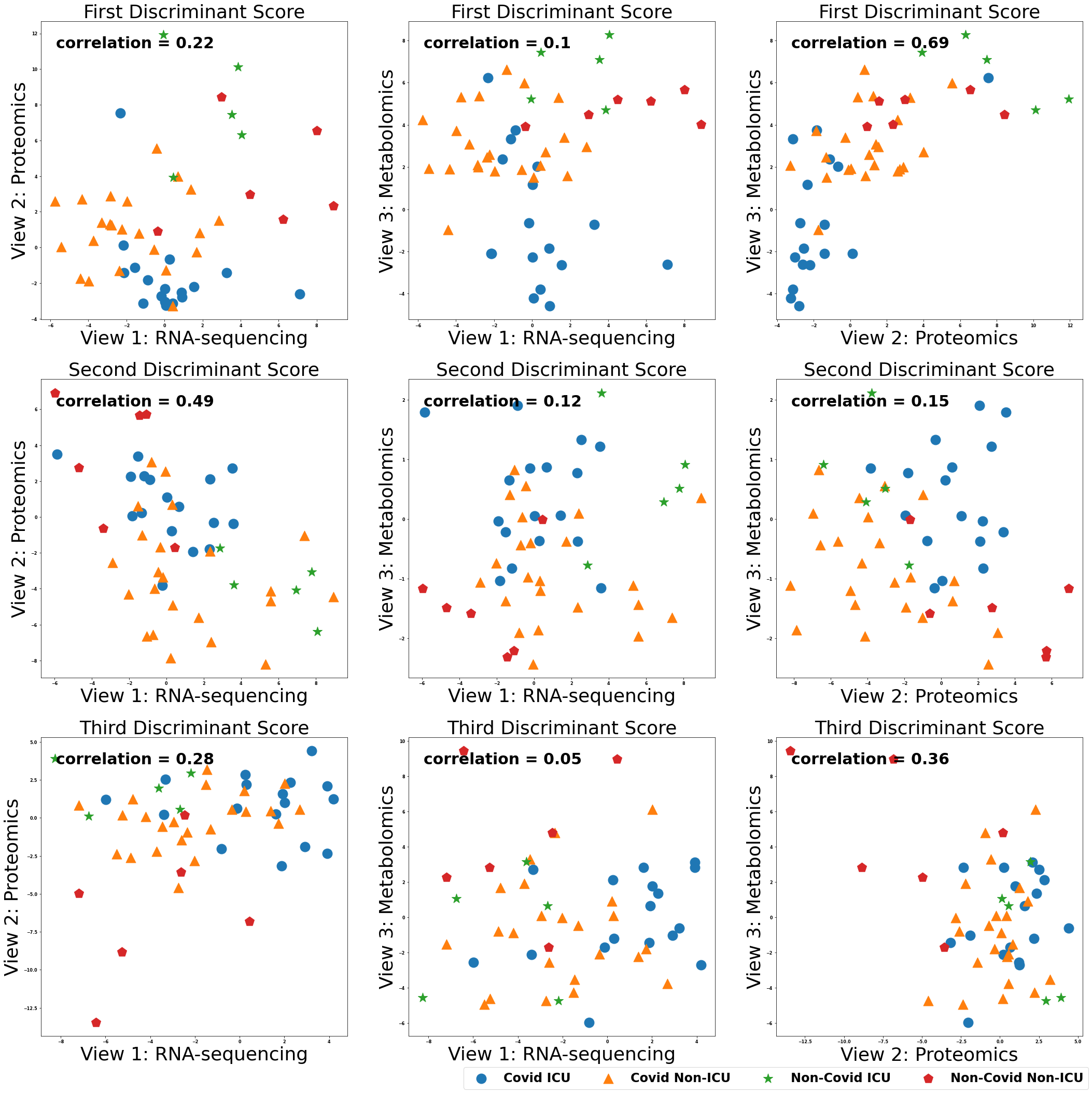}\\
	\end{tabular}
	\caption{Discrimination (3-D) plots: COVID-19 patient groups are well-separated in the training data. From the testing data, the COVID ICU group seems to be separated from the COVID NON-ICU and NON-COVID ICU groups, especially in the RNA-sequencing and proteomics data. Correlation plots (2-D): Overall (combining all three discriminant scores), the mean correlation between the metabolomics and proteomics data was highest (0.4) while the mean correlation between the metabolomics and RNA-sequencing data was lowest (0.09). }
	\label{fig:discorr}
\end{figure}

\begin{figure}
	\begin{tabular}{cc}
		\centering
		\includegraphics[width=0.45\textwidth]{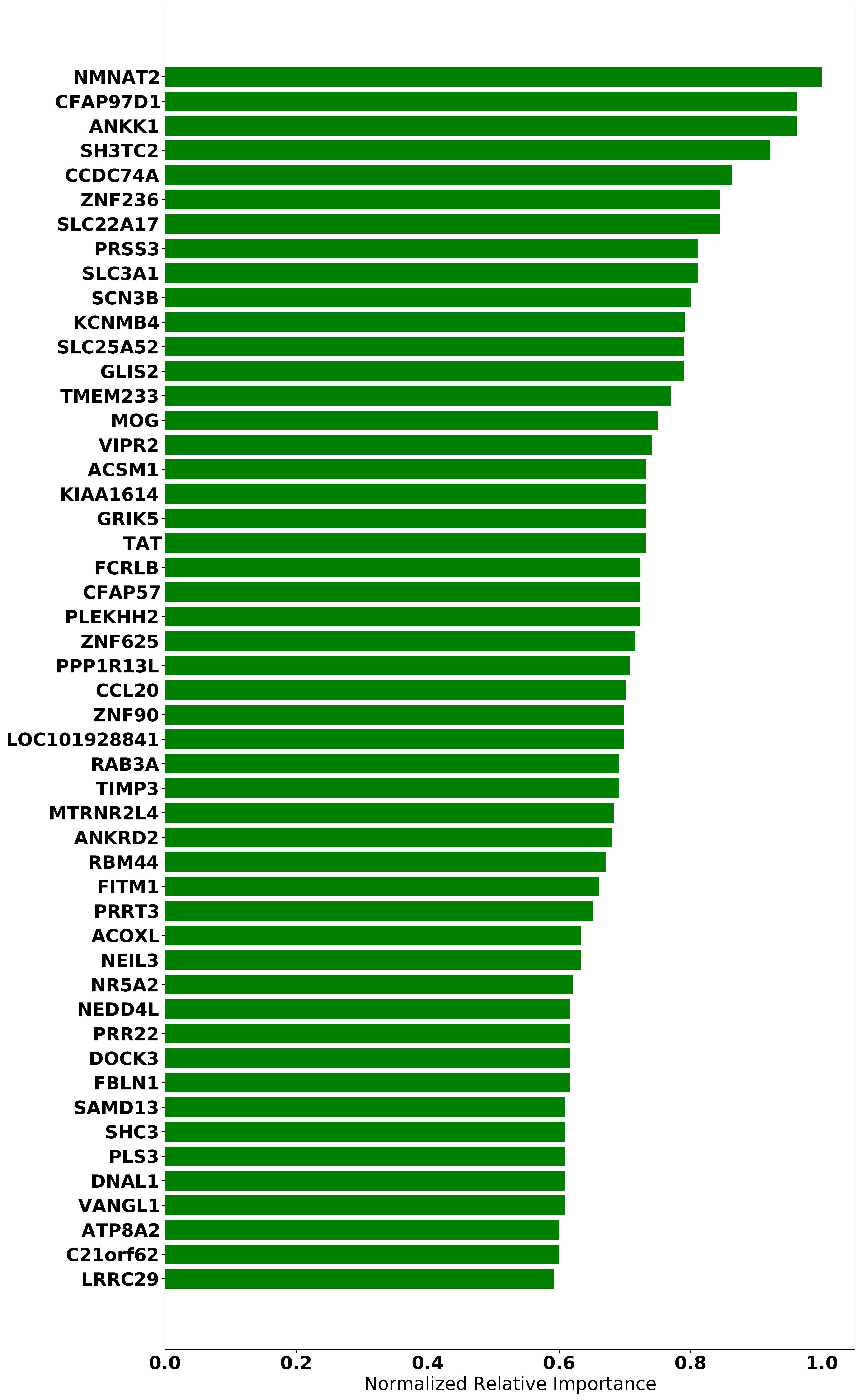}&
		\includegraphics[width=0.45\textwidth]{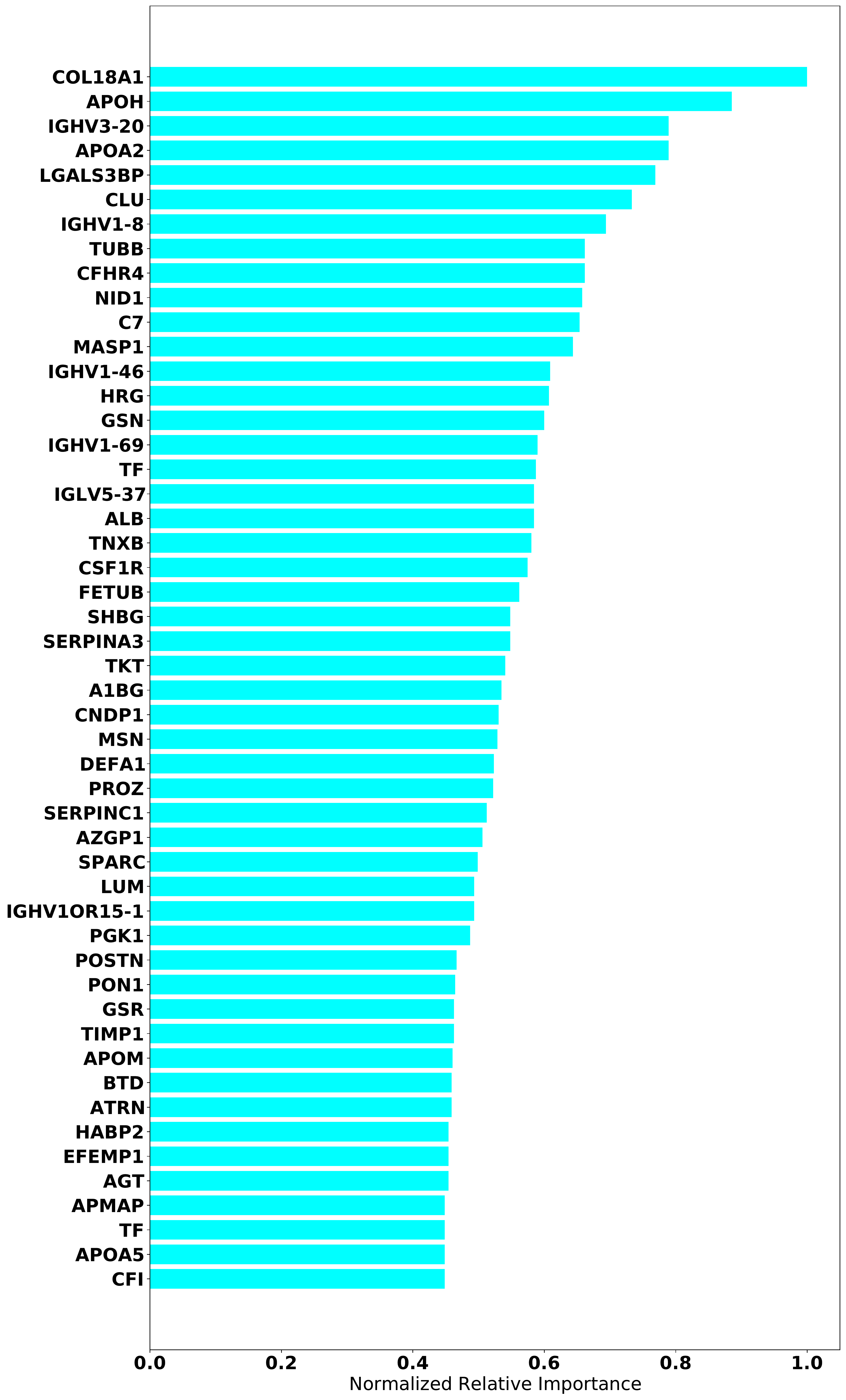}\\
		\includegraphics[width=0.50\textwidth]{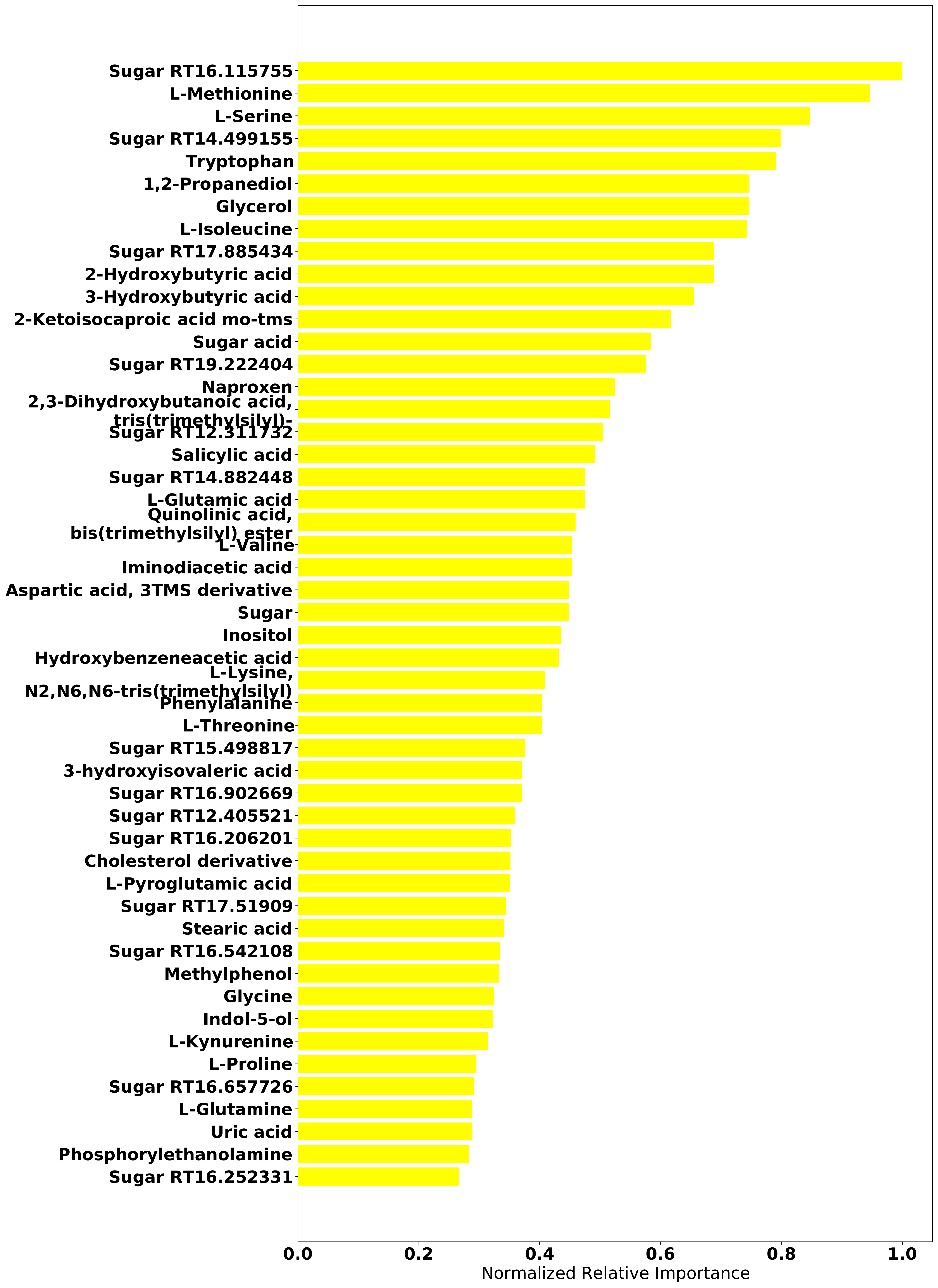} \\ 
	\end{tabular}
	\caption{Feature importance plots of the omics data used in the COVID-19 application. Upper left: RNA-Seq; upper right:  Proteomics ; lower left: Metabolomics.  Feature importance for each variable was normalized to the feature ranked highest for each omics. }
	\label{fig:featuresel}
\end{figure}

Table \ref{tab: omics} gives the test accuracy for Deep IDA in comparison with deep generalized CCA (Deep GCCA), SIDA, and SVM. 
Deep IDA on selected features and  Deep IDA refer to applying the proposed method with and without variable selection, respectively. Deep IDA with feature selection based on the top $10\%$  variables yield the same test classification accuracy as Deep IDA without feature selection, and these estimates are higher than the test accuracy from the other methods. Further, we observed a slight increase in classification performance when we implemented Deep IDA with feature selection based on the top 50 ranked molecules for each omics.   Naively stacking the data and applying  support vector machine results in the worse classification accuracy. When we applied SVM on the stacked data obtained from  the variables that were selected by Deep IDA, the  classification accuracy increased to $62.74\%$ (based on the top $10\%$ features selected by Deep IDA), representing a $13.73\%$ increase from applying SVM on the stacked data obtained from all the variables. We also observed an increase in classification accuracy when we implemented Deep GCCA  on the top $10\%$ selected features from Deep IDA compared to Deep GCCA  on the whole data. Compared to SIDA, the joint association and classification method that assesses linear relationships in the views and among the groups, the proposed method has a higher test accuracy. Figure \ref{fig:discorr} gives the discriminant and correlation plots from Deep IDA based on the top-ranked 50 molecules from each omics. From the discriminant plots of the first three discriminant scores, we notice that the samples are well-separated in the training data. For the testing data, we observe some overlaps in the sample groups but the COVID ICU group seems to be separated from the COVID NON-ICU and NON-COVID ICU groups. This separation is more apparent in the RNA-sequencing and proteomics data and less apparent in the metabolomics data. Further, based on the testing data, the correlation between the metabolomics and proteomics data was higher when considering the first and third discriminant scores (0.69 and 0.36, respectively). When considering the second discriminant score, the correlation between the RNA-sequencing and proteomics data was higher (0.49). Overall, the mean correlation between the metabolomics and proteomics data was highest (0.4) while the mean correlation between the metabolomics and RNA-sequencing data was lowest (0.09).
These findings suggest that the proposed method is capable of modeling nonlinear relationships among the different views and groups, and has potential to identify features that can lead to better classification results. Figure \ref{fig:featuresel} gives the top 50  genes, proteins, and metabolomics features that were highly-ranked by  Deep IDA. Feature importance for each variable was normalized to the feature ranked highest for each omics.  

\clearpage
\subsubsection{Pathway analysis of highly-ranked molecules}
\begin{table}
	\spacingset{1}
	\centering
	\caption{COVID-19 Omics data: SVM is based on stacked three-view raw data. DeepCCA+SVM is training SVM based on the last layer of DeepCCA. DeepSIDA applies nearest centroid  on the last layer for classification.	The reported classification accuracy for Deep IDA are based on optimized network structure. DeepIDA + top 10\% is based on input-512-20 network structure. DeepIDA + top 50 is based on input-512-256-20 network structure.\label{tab: omics}}
	\begin{tabular}{lr}
		\hline
		\hline
		Method&Accuracy (\%)  	\\
		\hline
		\hline
		SVM & 49.01\\
		SVM on selected top 10\% features & 62.74\\
		SVM on selected top 50 features & 64.71\\
		Deep GCCA + SVM & 64.71\\
		Deep GCCA + SVM on selected top 10\% features & 68.63\\
		Deep GCCA + SVM on selected top 50 features & 64.71\\
		SIDA & 60.78\\
		\rb{Deep IDA} & 76.47\\
		\rb{Deep IDA} on selected top 10\% features & 76.47\\
		\rb{Deep IDA} on selected top 50 features & 78.43\\
		\hline
		\hline
	\end{tabular}
\end{table}	
We used the Ingenuity Pathway Analysis (IPA) software  to investigate the molecular and cellular functions, pathways, and diseases  enriched in the proteins, genes, and metabolites that were ranked in the top 50  by our variable selection method. IPA searches the ingenuity pathway knowledge base, which is manually curated from scientific literature and over 30 databases, for gene interaction. We observed  strong pathways, molecular and cellular functions, and disease enrichment (Supplementary Tables 1 and 2). The top disease and disorders significantly enriched in our list of genes are found in Supplementary Table 2. We note that 36 of the biomolecules in our gene list were determined to be associated with neurological diseases. This finding aligns with studies that suggest that persons with COVID-19 are likely to have neurological manifestations such as reduced consciousness and stroke \cite{berlit2020neurological,taquet20216}. Further, 48 genes from our list were determined to be associated with cancer. Again, this supports studies that suggest that individuals with immuno-compromised system from cancer or individuals who recently recovered from cancer, for instance, are at higher risk for severe outcomes. Compared to the general population, individuals with cancer have a two-fold increased risk of contracting SARS-CoV-2\cite{al2020practical}.  The top 2 networks determined to be enriched in our gene list was ``hereditary disorder, neurological disease, organismal injury and abnormalities", and ``cell signaling, cellular function and maintenance, and small molecule biochemistry". 

\begin{figure}
	        \centering
\begin{tabular}{c}
        \includegraphics[height=5cm,width=17cm]{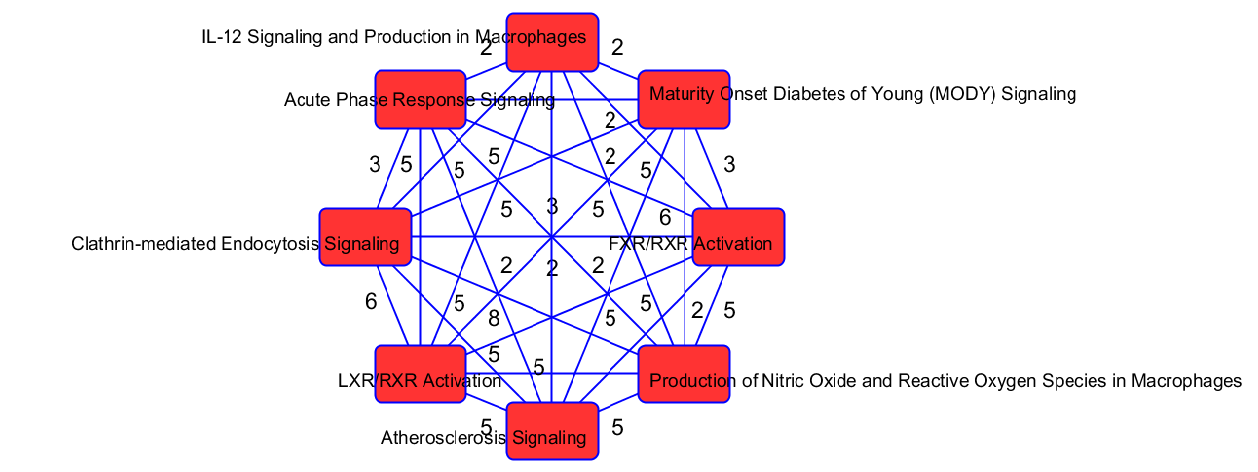}
\end{tabular}
 \caption{ Network of overlapping canonical pathways from highly ranked proteins.  Nodes refer to pathways and a line connects any two pathways when there is at least two  molecules in common between them. The two pathways ``FXR/RXR Activation” and ``LXR/RXR Activation” share a large number (eight) of molecules: AGT, ALB, APOA2, APOH, APOM, CLU, PON1 and TF. }\label{fig:protoverlap}
    \end{figure}
\begin{figure}
\centering
\begin{tabular}{c}
         \includegraphics[width=0.9\textwidth]{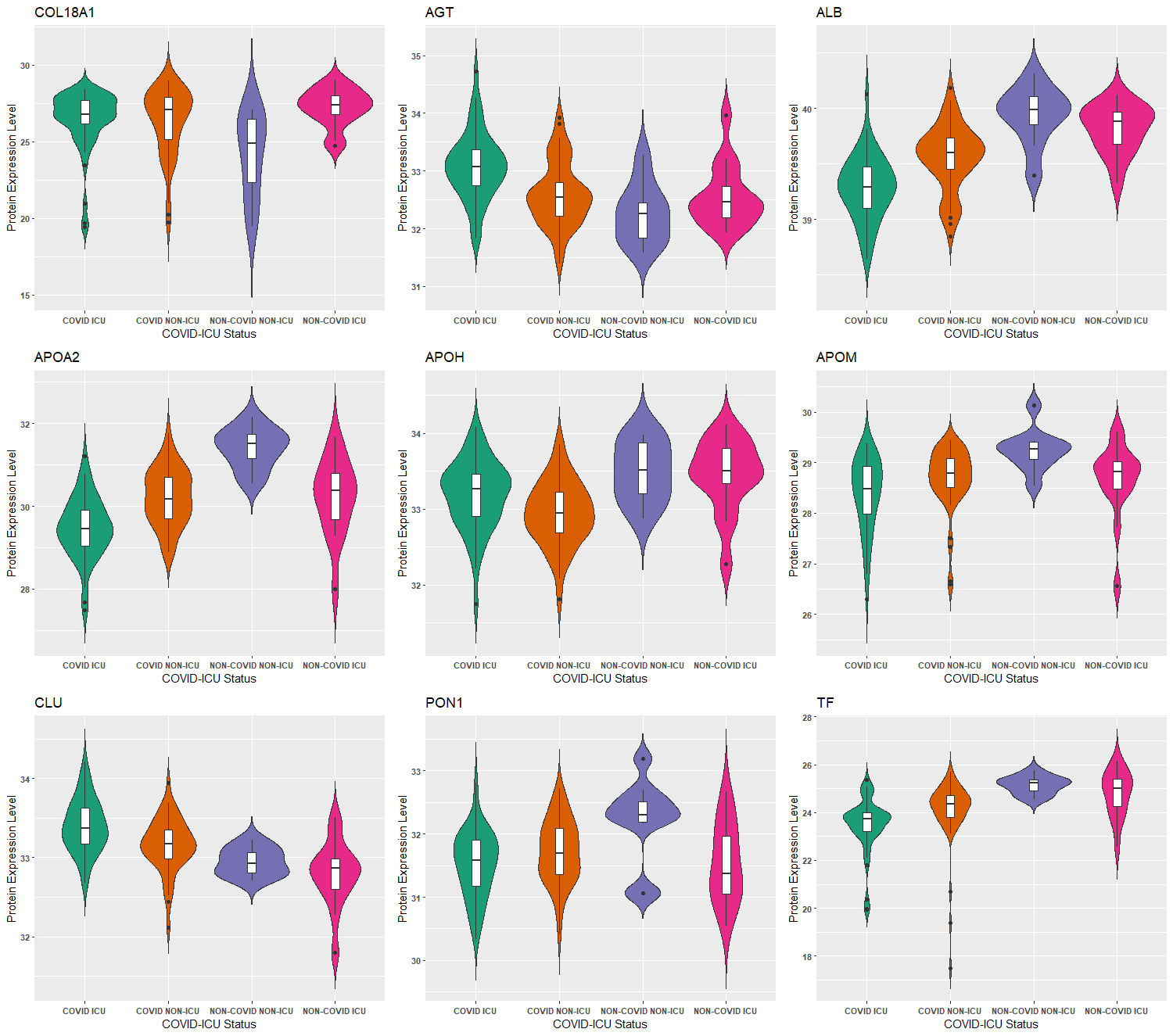}
 \end{tabular}
    \caption{Comparison of protein levels among COVID-19 patient groups (p-value $< 0.05$, Kruskal-Wallis test). COL18A1 was highly ranked by Deep IDA, and the other 8 proteins are shared by the ``FXR/RXR Activation” and ``LXR/RXR Activation” pathways. Protein expression levels for ALB, APOM, and TF  are lower in patients with COVID-19 (especially  in patients with COVID-19 who were admitted to the ICU). Protein expression levels for AGT and CLU are higher in patients with COVID-19 admitted to the ICU compared to the other groups.}
    \label{fig:proteinlevels}
\end{figure}

As in our gene list, 34 proteins were determined to be associated with neurological disease. Other significantly enriched diseases in our protein list included infectious diseases (such as  infection by SARS coronavirus), inflammatory response (such as inflammation of organ), and metabolic disease (including Alzheimer disease and diabetes mellitus).  A recent review \citep{steenblock2021covid} found that up to $50\%$ of those who have died from COVID-19 had metabolic and vascular disorders. In particular, patients with metabolic dysfunction (e.g., obesity, and diabetes) have an increased risk for developing severe COVID-19. Further, getting infected with SARS-CoV-2 can likely lead to new onset of diabetes.  The top 2 networks determined to be enriched in our protein list was ``infectious diseases, cellular compromise, inflammatory response", and ``tissue development, cardiovascular disease, hematological disease".  The top enriched canonical pathways in our protein list include the LXR/RXR activation FXR/RXR activation, acute phase response signaling, and atherosclerosis signaling (Table \ref{tab:toppathways}). These pathways are involved in metabolic processes such as cholesterol metabolism.  The molecular and cellular functions enriched in our protein list include cellular movement and lipid metabolism (Supplementary Table 2). Overlapping canonical pathways (Figure \ref{fig:protoverlap}) in IPA was used to visualize the shared biology in pathways through the common molecules participating in the pathways. 
The two pathways ``FXR/RXR Activation” and ``LXR/RXR Activation” share a large number (eight) of molecules: AGT, ALB, APOA2, APOH, APOM, CLU, PON1 and TF.  The LXR/RXR pathway is involved in the regulation of lipid metabolism, inflammation, and cholesterol to bile acid catabolism. The farnesoid X receptor (FXR) is a member of the nuclear family of receptors and plays a key role in metabolic pathways and regulating lipid metabolism, cell growth and malignancy \citep{wang2008fxr}.  We observed lower levels of ALB, APOM, and TF  in patients with COVID-19 (and more so in patients with COVID-19 who were admitted to the ICU)  relative to patients without COVID-19 (Figure \ref{fig:proteinlevels}). We also observed higher levels of AGT and CLU in patients with COVID-19 admitted to the ICU compared to the other groups. The fact that the top enriched pathways, and molecular and cellular functions are involved in metabolic processes such as lipid metabolism seem to  corroborate the findings in \citep{overmyer2021large} that  a key signature for COVID-19 is likely a dysregulated lipid transport system. 

For the top-ranked metabolomics features,  we first used the  MetaboAnalyst 5.0 \citep{metaboanalyst} software to obtain their human metabolome database reference ID  and then used IPA on the mapped data for pathway, diseases, molecular and cellular function enrichment analysis. Of the top 50 ranked features, we were able to map 25 features. 
The top disease and disorders significantly enriched in our list of metabolites ( Table 1 of the supplementary material) included cancer and gastrointestinal disease (such as digestive system, and hepatocellular cancer). Molecular and cellular functions enriched included amino acid metabolism, cell cycle, and cellular growth and proliferation. Figure 1 (supplementary material) shows the overlapping pathway network for the metabolites.  

Taken together, these findings suggest that COVID-19 disrupts many biological systems. The relationships found with diseases such as cancer, gastrointestinal, neurological conditions, and metabolic diseases (e.g., Alzheimers and diabetes mellitus) heighten the need to study the post sequelae effects of this disease in order to better  understand the mechanisms and to develop effective treatments. 

\begin{table}
\spacingset{1}
\small
 \caption{ Top 5 Canonical Pathways from Ingenuity Pathway Analysis (IPA).  }\label{tab:toppathways}
\begin{tabular}{|l|l|l|l|}
\hline
\textbf{Omics Data}               & \textbf{Top Canonical Pathway}                                                                                 & \textbf{P-value}  & \textbf{Molecules Selected}                                                                                                                                                                                \\ \hline
RNA Sequencing & 4-hydroxybensoate   Biosynthesis                                                                      & 2.07E-03 & TAT                                                                                                                                                                                               \\ \hline
               & 4-hydroxyphneylpyruvate   Biosythesis                                                                 & 2.07E-03 & TAT                                                                                                                                                                                               \\ \hline
               & Tyrosine   Degradation 1                                                                              & 1.03E-02 & TAT                                                                                                                                                                                               \\ \hline
               & Role   of IL-17A in Psoriasis                                                                         & 2.86E-02 & CCL20                                                                                                                                                                                             \\ \hline
               & Fatty   Acid Activation                                                                               & 2.86E-02 & ACSM1                                                                                                                                                                                             \\ \hline
Proteomics     & LXR/RXR   Activation                                                                                  & 4.14E-11 & \begin{tabular}[c]{@{}l@{}}AGT,   ALB, APOA2, APOH,  APOM, \\ CLU, PON1, TF\end{tabular}                                                                                                          \\ \hline
               & FXR/RXR   Activation                                                                                  & 5.02E-11 & \begin{tabular}[c]{@{}l@{}}AGT,   ALB, APOA2,  APOH, APOM, \\ CLU, PON1, TF\end{tabular}                                                                                                          \\ \hline
               & \begin{tabular}[c]{@{}l@{}}Acute   Phase Response \\ Signaling\end{tabular}                           & 3.30E-08 & \begin{tabular}[c]{@{}l@{}}AGT,   ALB, APOA2,  APOH, HRG, \\ SERPINA3, TF\end{tabular}                                                                                                            \\ \hline
               & Atherosclerosis   Signaling                                                                           & 1.06E-07 & \begin{tabular}[c]{@{}l@{}}ALB,   APOA2, APOM, CLU, \\ COL18A1, PON1\end{tabular}                                                                                                                 \\ \hline
               & \begin{tabular}[c]{@{}l@{}}Clathrin-mediated   Endocytosis \\ Signaling\end{tabular}                  & 1.09E-06 & \begin{tabular}[c]{@{}l@{}}ALB,   APOA2, APOM, CLU, \\ PON1, TF\end{tabular}                                                                                                                      \\ \hline
Metabolomics   & tRNA   Charging                                                                                       & 2.25E-13 & \begin{tabular}[c]{@{}l@{}}L-glutamic   acid,  L-Phenylalanine, \\ L-Glutamine, Glycine, L-Serine, \\ L-Methionine; L-Valine,  L-Isoleucine, \\ L-Threonine, L-Tryptophan, L-Proline\end{tabular} \\ \hline
               & Glutamate   Receptor Signaling                                                                        & 5.43E-05 & L-glutamic   acid,  glycine, L-Glutamine                                                                                                                                                          \\ \hline
               & \begin{tabular}[c]{@{}l@{}}Phenylalanine   Degradation IV \\ (Mammalian, via Side Chain)\end{tabular} & 7.24E-05 & \begin{tabular}[c]{@{}l@{}}L-glutamic   acid,  glycine, L-Glutamine, \\ L-Phenylalanine\end{tabular}                                                                                              \\ \hline
               & \begin{tabular}[c]{@{}l@{}}Superpathway   of Serine and \\ Clycine Biosynthesis I\end{tabular}        & 3.28E-04 & L-glutamic   acid,  glycine, L-serine                                                                                                                                                             \\ \hline
               & y-glutamyl   Cycle                                                                                    & 4.23E-04 & \begin{tabular}[c]{@{}l@{}}L-glutamic   acid,  glycine, \\ pyrrolidonecarboxylic acid\end{tabular}                                                                                                \\ \hline
\end{tabular}
\end{table}

\clearpage

\section{Conclusion}
We have proposed a deep learning method, Deep IDA,   for joint integrative analysis and classification studies of multi-view data.  Our framework extends the joint association and classification method proposed in \cite{safosida:2021} to model nonlinear relationships among multiple views and between classes in a view. Specifically, we use deep neural networks (DNN) to nonlinearly transform each view and we construct an optimization problem that takes
as input the output from our DNN and learns view-specific projections that result in maximum linear correlation of the transformed views and maximum linear separation within each view. Further, unlike most existing nonlinear  methods for integrating data from multiple views, we have  proposed a feature ranking approach based on resampling to identify features contributing most to the dependency structure among the views and the separation of classes within a view.  Our framework for feature ranking is general and applicable to other nonlinear methods for multi-view data integration. The
proposed algorithm, developed in Python 3, is  user-friendly and will be
useful in many data integration applications. Through simulation studies, we showed that the proposed
method outperforms several other linear and nonlinear methods for integrating data from multiple views, even in high-dimensional scenarios where the sample size is typically smaller than the number of variables.    

When Deep IDA was applied to proteomics, RNA sequencing, and metabolomics data obtained from individuals with and without COVID-19 who were or were not admitted to the ICU, we identified several molecules that better discriminated the COVID-19 patient groups. We also performed enrichment analysis of the molecules that were highly ranked and we observed strong pathways, molecular and
cellular functions, and disease enrichment. The top disease and disorders significantly enriched in our list of genes, proteins, and metabolomics data included cancer, neurological disorders, infectious diseases, and metabolic diseases.  While some of these findings corroborate earlier results, the top-ranked molecules could be further investigated to delineate their impact on COVID-19 status and severity. 

Our work has some limitations. First, the bootstrap technique proposed is computationally tasking. In our algorithm, we use parallelization to mitigate against the computational burden, however, more is needed to make the approach less expensive. Second, the proposed method has focused on binary or categorical outcomes. Future work could consider other outcome types (e.g., continuous and survival). Third, the number (or proportion) of top-ranked features need to be specified in advance. In our proposed bootstrap method, once we have identified the top-ranked variables, we fit another deep learning model to obtain low-dimensional representations of the data that result in maximum association among the views and separation of classes based on the top-ranked variables, and we use these to obtain test classification accuracy if testing data are available.  Alternatively, instead of learning a new model with the top-ranked variables, we could consider using the learned neural network parameters from the $M$ bootstrap implementations to construct $M$ $\bH^d_{test}$s, and then aggregate these (over $M$) to obtain an estimate of the view-specific top-level representations for classification. Future work could compare this alternative with the current approach. 

In conclusion, we have developed a deep learning method to jointly  model nonlinear relationships between data from multiple views and a binary or categorical outcome, while also producing highly-ranked features contributing most to the  overall association of the views and separation of the classes within  a view.  The encouraging simulations and real data findings, even for scenarios with small to moderate sample sizes,  motivate further applications. 


\section*{Funding and Acknowledgements}
The project described was supported by the Award Numbers 5KL2TR002492-04 from the National Center For Advancing Translational Science  and 1R35GM142695-01 from the National Institute Of General Medical Sciences of the National Institutes of Health.  The content is solely the responsibility of the authors and does not  represent the official views of the National Institutes of Health.

~\\
\textit{Declaration of Conflicting Interests}: The authors declare that there is no conflict of interest.

\section*{Data Availability and Software}
The data used were obtained from \cite{overmyer2021large}. 
We provide a Python package, \textit{Deep IDA}, to facilitate the use of our method. Its source  codes, along with a README file would be made available via \url{https://github.com/lasandrall/Deep IDA}. 

\section*{Supplementary Data}
In the online Supplementary Materials, we provide proof of Theorems 1 and 2, and also give more results from  real data analyses.

\end{document}


\def\spacingset#1{\renewcommand{\baselinestretch}%
{#1}\small\normalsize} \spacingset{1}

\title{Supplementary Material for ``Deep IDA: A Deep Learning Method for Integrative Discriminant Analysis of Multi-View Data with Feature Ranking--An Applicaition to COVID-19 severity"
}
\author{Jiuzhou Wang, Sandra E. Safo \\ Division of Biostatistics\\University of Minnesota, MN }
\date{}
\bigskip
\maketitle

\spacingset{1.0} 
\section{Theorems and Proofs}
\begin{thm}\label{thm:GEVPmain}
	Let $\bS_{t}^{d}$ and $\bS_{b}^{d}$ respectively be the total covariance and the between-class covariance for the top-level representations  $\bH^d, d=1,\ldots,D$. Let $\bS_{dj}$ be the cross-covariance between top-level representations $d$ and $j$.   Assume $\bS_{t}^{d} \succ 0$.  Define $\mathcal{M}^d = {\bS_t^d}^{-\frac{1}{2}} \bS_b^d {\bS_t^d}^{-\frac{1}{2}}$ and  $\mathcal{N}_{dj} = {\bS_t^d}^{-\frac{1}{2}} \bS_{dj} {\bS_t^j}^{-\frac{1}{2}}$. 
	Then $\bGamma^d \in \Re^{o_d \times l}$, $l \le \min\{K-1, o_1,\ldots,o_D\}$ in equation (4) of main text are eigenvectors  corresponding respectively to eigenvalues $\bLambda_{d}=$diag$(\lambda_{d_{k}},\ldots,\lambda_{d_{l}})$, $\lambda_{d_{k}} > \cdots > \lambda_{d_{l}}>0$ that iteratively solve the 
 eigensystem problems:
\begin{align*}
    \left(c_1\mathcal{M}^d + c_2\sum_{j\neq d}^D \mathcal{N}_{dj}\bGamma_j\bGamma_j^{\smt}\mathcal{N}_{dj}^{\smt}\right) \bGamma_d &= \bLambda_d \bGamma_d, \forall d = 1,...,D
\end{align*}
where $c_1 = \frac{\rho}{D}$ and $c_2 = \frac{2(1-\rho)}{D(D-1)}$. 
\end{thm}

\textbf{Prove}. Solving the optimization problem is equivalent to iteratively solving the  following generalized eigenvalue systems:
\begin{align*}
    \left(c_1\mathcal{M}^1 + c_2\sum_{j=2}^D \mathcal{N}_{1j}\bGamma_j\bGamma_j^T\mathcal{N}_{1j}^T\right) \bGamma_1 &= \bLambda_1 \bGamma_1\\
    &\vdots\\
    \left(c_1\mathcal{M}^d + c_2\sum_{j=1,j\neq d}^D \mathcal{N}_{dj}\bGamma_j\bGamma_j^T\mathcal{N}_{dj}^T\right) \bGamma_d &= \bLambda_d \bGamma_d\\
    &\vdots\\
    \left(c_1\mathcal{M}^D + c_2\sum_{j=1}^{D-1} \mathcal{N}_{Dj}\bGamma_j\bGamma_j^T\mathcal{N}_{Dj}^T\right) \bGamma_D &= \bLambda_D \bGamma_D
\end{align*}
where $c1 = \frac{\rho}{D}$ and $c2 = \frac{2(1-\rho)}{D(D-1)}$.\\
\begin{proof}
The Lagrangian is
\begin{small}
\begin{eqnarray}
    L(\bGamma_1,...,\bGamma_D,\lambda_1,...,\lambda_D)
    &=& \rho \frac{1}{D} \sum_{d=1}^D tr[\bGamma_d^T\mathcal{M}^d\bGamma_d] + (1-\rho) \frac{2}{D(D-1)} \sum_{d=1}^D \sum_{j,j\neq d}^D tr[\bGamma_d^T \mathcal{N}_{dj}  \bGamma_j \bGamma_j^T \mathcal{N}_{dj}^T \bGamma_d] - \sum_{d=1}^D \eta_d (tr[\bGamma_d^T\bGamma_d] - l)\nonumber\\
    &=& c_1 \sum_{d=1}^D tr[\bGamma_d^T\mathcal{M}^d\bGamma_d] + c_2 \sum_{d=1}^D \sum_{j,j\neq d}^D tr[\bGamma_d^T \mathcal{N}_{dj}  \bGamma_j \bGamma_j^T \mathcal{N}_{dj}^T \bGamma_d] - \sum_{d=1}^D \lambda_d (tr[\bGamma_d^T\bGamma_d] - l)
\end{eqnarray}
\end{small}
The first order stationary solution for $\bGamma_d (\forall d =1,...,D)$ is
\begin{equation}\label{eqn:partial_L}
    \frac{\partial L(\bGamma_1,...,\bGamma_D,\lambda_1,...,\lambda_D)}{\partial \bGamma_d^T}
    = 2c_1 \mathcal{M}^d \bGamma_d + 2c_2 \sum_{j,j\neq d}^D(\mathcal{N}_{dj}  \bGamma_j \bGamma_j^T \mathcal{N}_{dj}^T) \bGamma_d -2\lambda_d \bGamma_d = \textbf{0}
\end{equation}
Rearranging the equation \ref{eqn:partial_L} we have
\begin{align*}
    \left(c_1\mathcal{M}^d + c_2\sum_{j,j\neq d}^D \mathcal{N}_{dj}\bGamma_j\bGamma_j^T\mathcal{N}_{dj}^T\right) \bGamma_d &= \lambda_d \bGamma_d
\end{align*}
For $\bGamma_j,\forall j\neq d$ fixed, the above can be solved for the eigenvalues of $(c_1\mathcal{M}^d + c_2\sum_{j,j\neq d}^D \mathcal{N}_{dj}\bGamma_j\bGamma_j^T\mathcal{N}_{dj}^T)$. Arrange the eigenvalues from large to small and denote $\bLambda_d \in \bR^{o_d\times o_d}$ as the diagonal matrix of those values. For the top $l$ largest eigenvalues, denote the corresponding eigenvectors as $\widehat{\bGamma}_d = [\gamma_{d,1},...,\gamma_{d,l}]$. Therefore, starting from $d=1$, following this process, $\widehat{\bGamma}_1$ is updated; then, update $\widehat{\bGamma}_2$ and so on; finally, update $\widehat{\bGamma}_D$. We iterate until convergence, which is defined as,  $\frac{\|\widehat{\bGamma}_{d_{,new}} - \widehat{\bGamma}_{d_{,old}}\|_F^2}{\| \widehat{\bGamma}_{d_{,old}}\|_F^2} < \epsilon$. When convergence is achieved, set $\widetilde{\bGamma}_d=\widehat{\bGamma}_d, \forall d=1,...,D$.
\end{proof}

\begin{thm}\label{thm:loss}
For $d$ fixed, let $\eta_{d,1}, \ldots, \eta_{d,l}$, $l \le \min\{K-1, o_1,\dots,o_D\}$ be the largest $l$ eigenvalues of $c_1\mathcal{M}^d + c_2\sum_{j\neq d}^D \mathcal{N}_{dj}\bGamma_j\bGamma_j^{\smt}\mathcal{N}_{dj}^{\smt}$. Then the solution $\widetilde{f}^d$ to the optimization problem in equation (5) [main text] for view $d$ maximizes
\begin{align}
    \sum_{r=1}^l \eta_{d,r}.
\end{align}
\end{thm}

\textbf{Proof}. Fix $d$ and let $\eta_{d,1},\eta_{d,2},...,\eta_{d,l}$ be the top $l$ eigenvalues of 
\begin{align*}
    c_1\mathcal{M}^d + c_2\sum_{j\neq d}^D \mathcal{N}_{dj}\widetilde{\bGamma}_j\widetilde{\bGamma}_j^T\mathcal{N}_{dj}^T.
\end{align*}
Then, 
\begin{align*}
    \sum_{r=1}^l \eta_{d,r} = c_1  tr[\widetilde{\bGamma}_d^T\mathcal{M}^d\widetilde{\bGamma}_d] + c_2 \sum_{j,j\neq d}^D tr[\widetilde{\bGamma}_d^T \mathcal{N}_{dj}  \widetilde{\bGamma}_j \widetilde{\bGamma}_j^T \mathcal{N}_{dj}^T \widetilde{\bGamma}_d]
\end{align*}
\begin{proof}
\begin{align*}
    &c_1  tr[\widetilde{\bGamma}_d^T\mathcal{M}^d\widetilde{\bGamma}_d] + c_2 \sum_{j,j\neq d}^D tr[\widetilde{\bGamma}_d^T \mathcal{N}_{dj}  \widetilde{\bGamma}_j \widetilde{\bGamma}_j^T \mathcal{N}_{dj}^T \widetilde{\bGamma}_d]\\
    &=  tr(\widetilde{\bGamma}_d^T (c_1\mathcal{M}^d + c_2\sum_{j,j\neq d}^D \mathcal{N}_{dj}\widetilde{\bGamma}_j\widetilde{\bGamma}_j^T\mathcal{N}_{dj}^T) \widetilde{\bGamma}_d)\\
    &= tr(\widetilde{\bGamma}_d^T \bLambda_d \widetilde{\bGamma}_d)\\
    &=  \sum_{r=1}^l \eta_{d,r}
\end{align*}
\end{proof}

\section{More Results From Real Data Analysis}
\subsection{Data preprocessing and application of Deep IDA and competing methods}
Of the 128 patients, 125 had both omics and clinical data. We focused on proteomics, RNA-seq, and metabolomics data in our analyses since many lipids were not annotated.  We formed a four-class classification problem using COVID-19 and ICU status. Our four groups were: with COVID-19 and not admitted to the ICU (COVID Non-ICU), with COVID-19 and admitted to the ICU (COVID ICU), no COVID-19 and admited to the ICU (Non-COVID ICU), and no COVID-19 and not admitted to the ICU (Non-COVID Non-ICU). The frequency distribution of samples in these four groups were: $40\%$ COVID ICU, $40\%$ COVID Non-ICU, $8\%$ Non-COVID Non-ICU, and  $12\%$ Non-COVID ICU. The initial dataset contains 18,212 genes, 517 proteins, and 111 metabolomics features. Prior to applying our method, we pre-processed the data as follows. 
All genes which were missing in our samples were removed from the dataset and 15,106 genes remained. We selected genes that more than half of their variables are non-zero, and we applied box-cox transformation on each gene as the gene data were highly skewed. The transformed data were standardized to have mean zero and variance one. We kept genes with variance less than the 25th percentile. We then used ANOVA on the standardized data to filter out (p-values $> 0.05$) genes with low potential to discriminate among the four groups.  For the proteomics and metabolomics data, we standardized each molecule to have mean zero and variance one, pre-screened with ANOVA and filtered out molecules with p-values $> 0.05$. Our final data were $\mathbf{X}^{1} \in \Re^{125 \times 2,734}$ for the gene data, $\mathbf{X}^{2} \in \Re^{125 \times 269}$ for the protoemics data, and $\mathbf{X}^{3} \in \Re^{125 \times 66}$ for the metabolomics data. 

\begin{figure}[H]
\begin{tabular}{c}
        \centering
       \includegraphics[height=6cm,width=13cm]{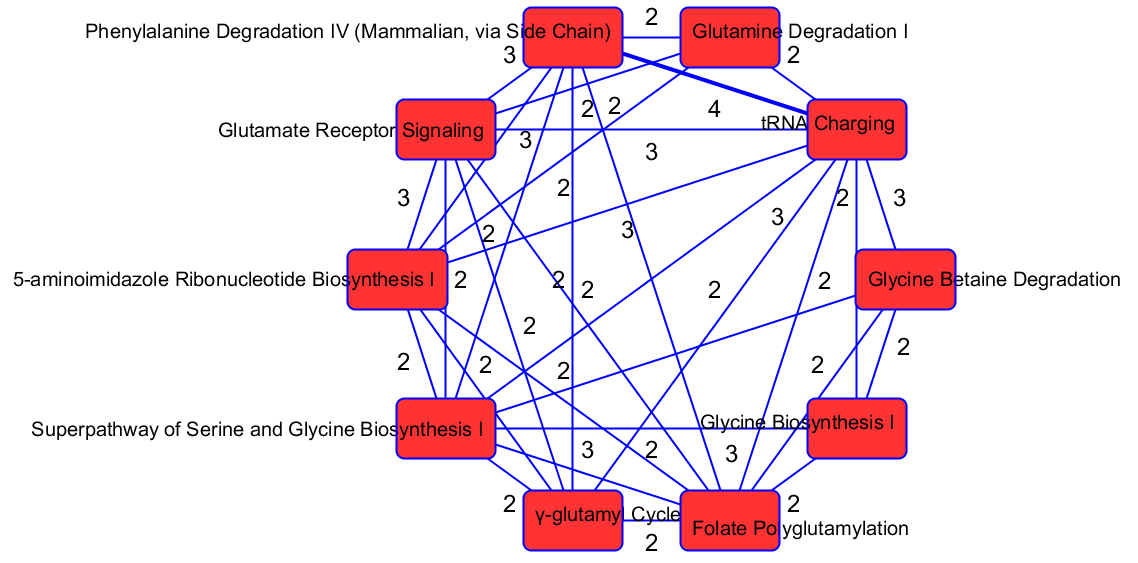}
\end{tabular}
 \caption{ Network of overlapping canonical pathways from highly ranked metabolites.  Nodes refer to pathways and a line connects any two pathways when there is at least two  molecules in common between them. }\label{fig:metabolitesoverlap}
    \end{figure}

\begin{table}[H]
 \caption{ Top Diseases and Biological Functions from Ingenuity Pathway Analysis (IPA).  }\label{tab:topdiseases}
\begin{tabular}{|l|l|l|l|}
\hline
               & Top Diseases and Bio Functions                                                                                                                       & P-value range         & Molecules Selected \\ \hline
RNA Sequencing & \begin{tabular}[c]{@{}l@{}}Cancer   (such as non-melanoma solid tumor, \\ head and neck tumor)\end{tabular}                                          & 4.96E-02   – 2.74E-05 & 48                 \\ \hline
               & Organismal   Injury and Abnormalities                                                                                                                & 4.96E-02   – 2.74E-05 & 48                 \\ \hline
               & \begin{tabular}[c]{@{}l@{}}Neurological   Disease (such as glioma cancer, \\ brain lesion, neurological deficiency)\end{tabular}                     & 4.86E-02   – 5.84E-05 & 36                 \\ \hline
               & \begin{tabular}[c]{@{}l@{}}Developmental   Disorder (such as intellectual \\ diability with ataxia)\end{tabular}                                     & 4.46E-02   – 3.76E-05 & 16                 \\ \hline
               & Hereditary   Disorder (such as familial midline effect)                                                                                              & 4.86E-02   – 2.02E-05 & 16                 \\ \hline
               &                                                                                                                                                      &                       &                    \\ \hline
Proteomics     & \begin{tabular}[c]{@{}l@{}}Infectious   Diseases (such as Severe COVID-19, \\ COVID-19, infection by SARS coronavirus)\end{tabular}                  & 1.75E-03   – 8.31E-13 & 19                 \\ \hline
               & \begin{tabular}[c]{@{}l@{}}Inflammatory   Response (such as inflammation of \\ organ, degranulation of blood platelets)\end{tabular}                 & 1.29E-03   – 1.34E-12 & 32                 \\ \hline
               & \begin{tabular}[c]{@{}l@{}}Metabolic   Disease (such as amyloidosis, \\ Alzheimer disease, diabetes mellitus)\end{tabular}                           & 1.47E-03   – 2.48E-12 & 20                 \\ \hline
               & \begin{tabular}[c]{@{}l@{}}Organismal   Injury and Abnormalities \\ (such as amyloidosis, tauopathy)\end{tabular}                                    & 1.72E-03   – 2.48E-12 & 39                 \\ \hline
               & \begin{tabular}[c]{@{}l@{}}Neurological   Disease (such as tauopathy, progressive \\ encephalopathy, progressive neurological disorder)\end{tabular} & 1.57E-03   – 3.41E-11 & 34                 \\ \hline
               &                                                                                                                                                      &                       &                    \\ \hline
Metabolomics   & Cancer                                                                                                                                               & 3.63E-02   – 5.20E-14 & 18                 \\ \hline
               & \begin{tabular}[c]{@{}l@{}}Gastrointestinal   Disease (such as digestive system \\ cancer, hepatocellular carcinoma)\end{tabular}                    & 3.64E-02   – 5.20E-14 & 20                 \\ \hline
               & \begin{tabular}[c]{@{}l@{}}Organismal   Injury and Abnormalities \\ (such as digestive system cancer, abdominal cancer)\end{tabular}                 & 3.79E-02   – 5.20E-14 & 22                 \\ \hline
               & \begin{tabular}[c]{@{}l@{}}Hepatic   System Disease \\ (such as hepatocellular carcinoma, liver lesion)\end{tabular}                                 & 2.91E-02   – 1.66E-11 & 15                 \\ \hline
               & \begin{tabular}[c]{@{}l@{}}Developmental   Disorder \\ (such as mucopolysaccharidosis type I, spina bifida)\end{tabular}                             & 2.44E-02   – 1.83E-09 & 11                 \\ \hline
\end{tabular}
\end{table}

\newpage
\begin{table}
	\caption{ Top Molecular and Cellular Functions Functions from Ingenuity Pathway Analysis (IPA).  }\label{tab:topfunctions}
\begin{tabular}{|l|l|l|l|}
\hline
                 & Molecular   and Cellular Functions       & P-value   range       & Molecules   Selected \\ \hline
RNA   Sequencing & Cell   Death and Survival                & 4.46E-02   – 2.00E-03 & 8                    \\ \hline
                 & Amino   Acid Metabolism                  & 3.47E-02   – 2.07E-05 & 2                    \\ \hline
                 & Cell-to-cell   Signaling and Interaction & 4.86E-02   – 2.07E-03 & 10                   \\ \hline
                 & Cellular   Assembly and Organization     & 4.46E-02   – 2.07E-03 & 9                    \\ \hline
                 & Cellular   Function and Maintenance      & 4.86E-02   – 2.07E-03 & 10                   \\ \hline
                 &                                          &                       &                      \\ \hline
Proteomics       & Cellular   Compromise                    & 1.29E-03   – 1.34E-12 & 13                   \\ \hline
                 & Cellular   Movement                      & 1.65E-03   – 2.19E-09 & 24                   \\ \hline
                 & Lipid   Metabolism                       & 1.28E-03   – 2.95E-09 & 15                   \\ \hline
                 & Molecular   Transport                    & 1.28E-03   – 2.95E-09 & 15                   \\ \hline
                 & Small   molecule Biochemistry            & 1.61E-03   – 2.95E-09 & 19                   \\ \hline
                 &                                          &                       &                      \\ \hline
Metabolomics     & Amino   Acid Metabolism                  & 3.64E-02   – 3.99E-08 & 9                    \\ \hline
                 & Molecular   Transport                    & 3.82E-02   – 3.99E-08 & 17                   \\ \hline
                 & Small   Molecule Biochemistry            & 3.64E-02   – 3.99E-08 & 18                   \\ \hline
                 & Cellular   Growth and Proliferation      & 3.79E-02   – 5.12E-08 & 16                   \\ \hline
                 & Cell   Cycle                             & 3.63E-02   – 5.81E-07 & 10                   \\ \hline
\end{tabular}
\end{table}

\begin{table}
	\begin{small}
		\begin{centering}
			\caption{\textbf{Linear Simulations} Network structures for all deep learning based methods. In order to make fair comparisons, for each dataset, the network structure for Deep CCA/Deep GCCA is the same as the proposed Deep IDA method. The activation function is Leakly Relu with parameter 0.1 by default. After activation, batch normalization is also implemented. $-$ indicates not applicable}	\label{tab: NetworkStructure}
			\begin{tabular}{lllllll}
				\hline
				\hline
				Data & Sample size  & Feature size & Method & Network structure & Epochs  & Batch 	\\
				~& (Train,Valid,Test)  & ($p^1, p^2, p^3)$ & ~ & ~ &  per run &  size	\\
				
				\hline
				\hline
				Setting 1 & 540,540,1080 & 1000,1000,- & Deep IDA (+Bootstrap)	&Input-512-256-64-10& 50 & 540\\
				\hline
				Setting 1 & 540,540,1080 & 1000,1000,- & Deep CCA	&Input-512-256-64-10& 50 & 180\\
				\hline
				Setting 2 & 540,540,1080 & 1000,1000,1000 & Deep IDA (+Bootstrap)	&Input-512-256-20& 50 & 540\\
				\hline
				Setting 2 & 540,540,1080 & 1000,1000,1000 & Deep GCCA	&Input-512-256-64-20& 200 & 540\\
				
				\hline
				\hline
			\end{tabular}
		\end{centering}
	\end{small}
\end{table}

\begin{table}
	\begin{centering}
		\begin{scriptsize}	
			\caption{\textbf{Nonlinear Simulations} Network structures for all deep learning based methods. In order to make fair comparisons, for each dataset, the network structure for Deep CCA/Deep GCCA is the same as the proposed Deep IDA method. The activation function is Leakly Relu with parameter 0.1 by default. After activation, batch normalization is also implemented.}	\label{tab: NetworkStructure2}
			\begin{tabular}{lllllll}
				\hline
				\hline
				
				Data & Sample size  & Feature size & Method & Network structure & Epochs  & Batch 	\\
				~& (Train,Valid,Test)  & ($p^1, p^2, p^3)$ & ~ & ~ &  per run &  size	\\
				\hline
				\hline				
				Setting 1 & 350,350,350 & 500,500 & Deep IDA (+Bootstrap)	
				&Input-256*10-64-20& 50 & 350\\
				\hline
				Setting 1 & 350,350,350 & 500,500 & Deep CCA	
				&Input-256*10-64-20& 50 & 350\\
				\hline
				Setting 3b & 5250,5250,5250 & 500,500 & Deep IDA (+Bootstrap)	&Input-256-256-256-256-256-256-64-20& 50 & 500\\
				\hline
				Setting 3b & 5250,5250,5250 & 500,500 & Deep CCA	&Input-256-256-256-256-256-256-64-20& 50 & 500\\
				\hline
				Setting 4 & 350,350,350 & 2000,2000 & Deep IDA (+Bootstrap)	&  input-256-256-256-256-256-256-256-64-20& 50 & 350\\
				\hline
				Setting 4 & 350,350,350 & 2000,2000 & Deep CCA	&  input-256-256-256-256-256-256-256-64-20& 50 & 350\\
				\hline
				Setting 5b & 5250,5250,5250 & 2000,2000 & Deep IDA (+Bootstrap)	&Input-256-256-256-256-256-256-64-20& 50 & 500\\
				\hline
				Setting 5b & 5250,5250,5250 & 2000,2000 & Deep CCA	&Input-256-256-256-256-256-256-64-20& 50 & 500\\
				\hline
				\hline
			\end{tabular}
		\end{scriptsize}
	\end{centering}
\end{table}	

\begin{table}
	\begin{centering}
		\caption{\textbf{Real Data Analysis} Network structures for all deep learning based methods. In order to make fair comparisons, for each dataset, the network structure for Deep CCA/Deep GCCA is the same as the proposed Deep IDA method. The activation function is Leakly Relu with parameter 0.1 by default. After activation, batch normalization is also implemented. For Covid-19 data, we select the top 50 features for each view  from Bootstrap Deep IDA with input-512-20. $-$ indicates not applicable}	\label{tab: NetworkStructure3}
		\begin{tabular}{lllllll}
			\hline
			\hline
			Data & Sample size  & Feature size & Method & Network structure & Epochs  & Batch 	\\
			~& (Train,Valid,Test)  & ($p^1, p^2,p^3)$ & ~ & ~ &  per run &  size	\\
			\hline
			\hline
			\hline
			\hline
			Noisy  & 50000,10000, & 784,784,- & Deep CCA	&Input-512-256-64-20& 50 & 50000\\
			MNIST & 10000& ~ & non-Bootstrap Deep IDA	&~& ~ & ~\\
			
			\hline
			Covid-19  & 74,0,21 & 2734,269,66 & non-Bootstrap Deep IDA	&Input-512-20& 20 & 74\\
			\hline
			Covid-19  & 74,0,21 & 2734,269,66& Deep IDA on selected 	&Input-512-256-20& 20 & 74\\
			~ & ~ & ~ & top 50 features	&~& ~ &~\\
			\hline
			Covid-19  & 74,0,21 & 2734,269,66 & Deep IDA on selected 	&Input-256-64-20& 20 & 74\\
			~  & ~ & ~ &  top 10 percent features	&~& ~& ~\\
			\hline
			Covid-19 & 74,0,21 & 2734,269,66 &  Deep GCCA on 	&Input-256-20& 150 & 74\\
			~ & ~ & ~&  selected top 50 features	&~& ~ & ~\\
			\hline
			\hline
		\end{tabular}
	\end{centering}
\end{table}